\documentclass[conference, final]{IEEEtran}

\IEEEoverridecommandlockouts

\usepackage{setspace}
\usepackage{enumerate}
\usepackage{cite}
\usepackage{times}
\usepackage{url}
\usepackage{graphicx}
\usepackage{subfigure}
\usepackage{amsmath}
\usepackage{url}
\usepackage{color}
\usepackage{times}
\usepackage{xcolor}
\usepackage{latexsym}
\usepackage[utf8]{inputenc}
\usepackage[small]{caption}
\usepackage{graphicx}
\usepackage{booktabs}
\usepackage{multirow}
\usepackage{mathrsfs}
\usepackage{amsmath}
\usepackage{hyperref}
\usepackage{amssymb}
\usepackage{balance}

\newcommand{\refeqn}[1]{Equation \ref{#1}}
\newcommand{\reffig}[1]{Figure \ref{#1}}
\newcommand{\reftbl}[1]{Table \ref{#1}}
\newcommand{\refsec}[1]{Section \ref{#1}}

\newcommand{\m}[1]{\mathcal{#1}}
\newcommand{\method}{NeuralDater}
\newcommand{\methodnew}{Attentive NeuralDater}
\newcommand{\methodc}{Attentive Context Model}
\newcommand{\methode}{Ordered Event Model}
\newcommand{\methodfin}{AD3}
\newcommand{\methodkg}{HyTE}
\newcommand{\methodcs}{AC-GCN}
\newcommand{\methodes}{OE-GCN}

\newcommand*{\Scale}[2][4]{\scalebox{#1}{$#2$}}%
\begin{document}
\title{Timestamping Documents and Beliefs}

\author{
\IEEEauthorblockN{
Swayambhu Nath Ray
}
\IEEEauthorblockA{Department~of~Computational~and~Data~Sciences\\
Indian Institute of Science, Bangalore, India\\
swayambhuray@iisc.ac.in}
}

\maketitle

\begin{abstract}

Most of the textual information available to us are temporally variable. In a world where information is dynamic, time-stamping them is a very important task. Documents are a good source of information and are used for many tasks like, sentiment analysis, classification of reviews etc. The knowledge of creation date of documents facilitates several tasks like summarization, event extraction, temporally focused information extraction etc. Unfortunately, for most of the documents on the web, the time-stamp meta-data is either erroneous or missing. Thus document dating is a challenging problem which requires inference over the temporal structure of the document alongside the contextual information of the document. Prior document dating systems have largely relied on handcrafted features while ignoring such document-internal structures. In this paper we propose \method{}, a Graph Convolutional Network (GCN) based document dating approach which jointly exploits syntactic and temporal graph structures of document in a principled way. We also pointed out some limitations of \method{} and tried to utilize both context and temporal information in documents in a more flexible and intuitive manner proposing \methodfin{}: Attentive Deep Document Dater, an attention-based document dating system. To the best of our knowledge these are the first application of deep learning methods for the task. Through extensive experiments on real-world datasets, we find that our models significantly outperforms state-of-the-art baselines by a significant margin.

Relational facts or beliefs, are a good source of world knowledge. Knowledge graphs (KGs) encode these factual beliefs in form of triple (entity, relation, entity), e.g., \textit{(Brussels, isCapitalOf, Belgium)}.  KGs often show temporal dynamics, e.g., the fact \textit{(Cristiano\_Ronaldo, playsFor, Manchester\_United)} is valid only from 2003 to 2009. Hence time-stamping KG beliefs is an important task, which helps in answering queries related to time directly from KGs. In this paper we looked into temporally guided link prediction task using temporally rich knowledge graph embeddings. Most of the existing KG embedding methods ignore this temporal dimension while learning embeddings of the KG elements. We propose \methodkg{}: Hyperplane-based Temporally aware Knowledge Graph Embedding, a temporally aware KG embedding method which explicitly incorporates time in the entity-relation space by stitching each timestamp with a corresponding hyperplane. Through extensive experimentation on temporal datasets extracted from real world KGs, we demonstrate the effectiveness of our model over both traditional as well as temporal KG embedding methods.

We have explored these two related tasks in \refsec{sec:Document} and \refsec{sec:Knowledge} respectively.

\end{abstract}
\section{\textbf{Document Time-Stamping}}
\label{sec:Document}
\subsection{\textbf{Introduction}}
\label{sec:introduction}

{\bf{Document Time-Stamping:}}
Date of a document, also referred to as the Document Creation Time (DCT), is at the core of many important tasks, such as, information retrieval \cite{ir_time_usenix, ir_time_li,ir_time_dakka}, temporal reasoning \cite{temp_reasoner1,temp_reasoner2}, text summarization \cite{text_summ_time}, event detection \cite{event_detection}, and analysis of historical text \cite{history_time}, among others. In all such tasks, the document date is assumed to be available and also accurate -- a strong assumption, especially for arbitrary documents from the Web. Thus, there is a need to automatically predict the date of a document based on its content. This problem is referred to as \emph{Document Dating}.

\begin{figure}[t]
	\begin{minipage}{1in}
		\includegraphics[scale=0.58]{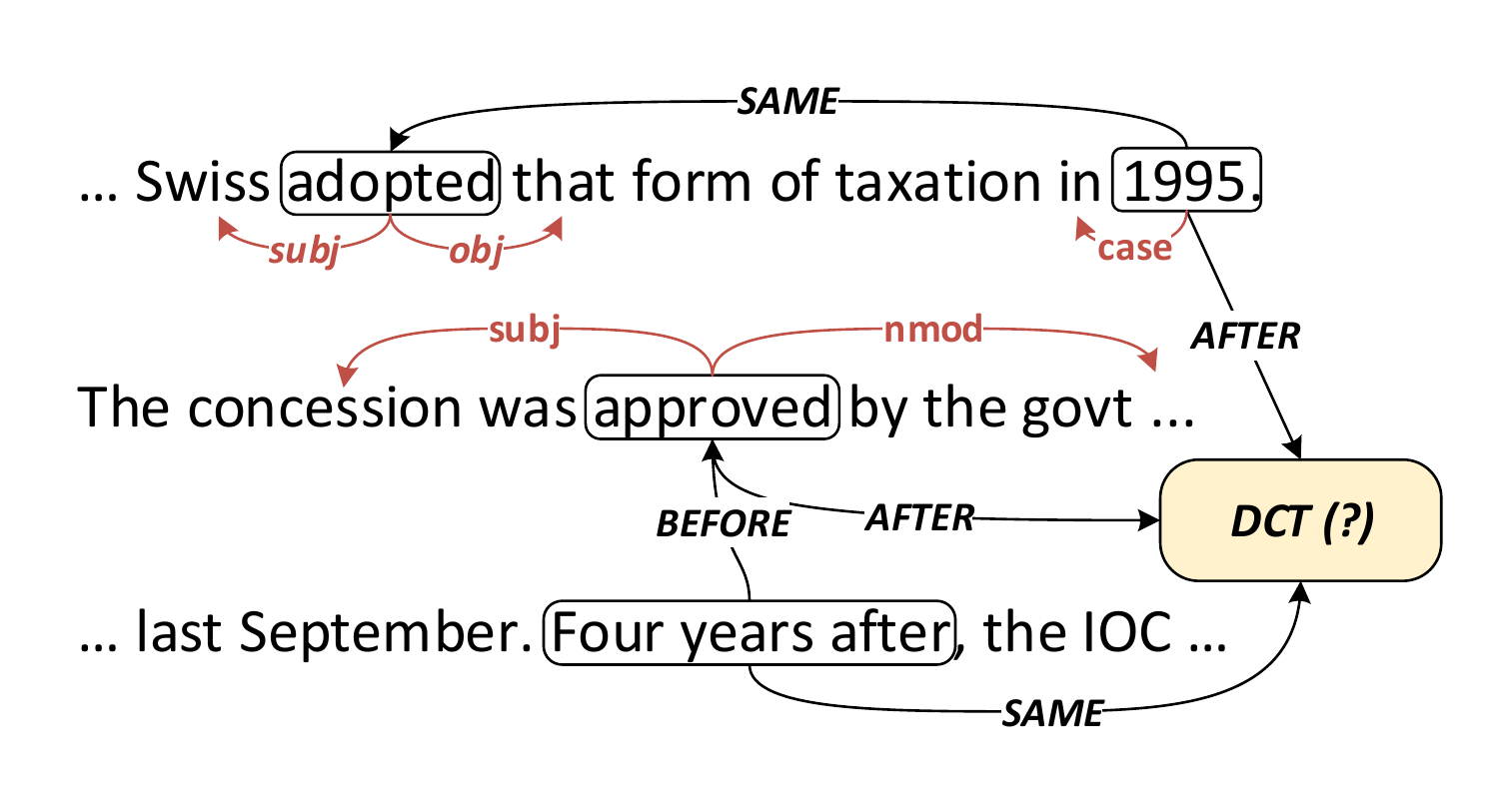}
	\end{minipage} \\
	\begin{minipage}{1in}
		\includegraphics[width=3.5in]{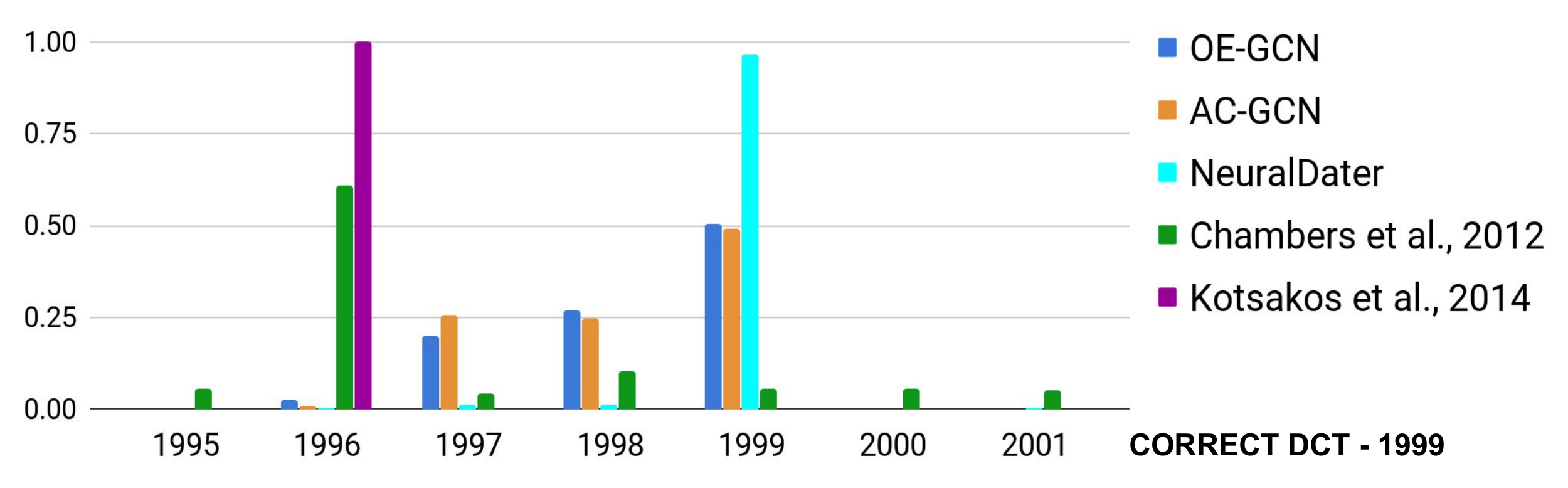}
	\end{minipage}
	\caption{\label{fig:motivation}\textbf{Top:} An example document annotated with syntactic and temporal dependencies. In order to predict the right value of 1999 for the Document Creation Time (DCT), inference over these document structures is necessary. \textbf{Bottom:} Document date prediction by two state-of-the-art-baselines and our models, the methods proposed in this paper. While the two previous methods are getting misled by the temporal expression (\textit{1995}) in the document, our models are able to use the syntactic and temporal structure of the document to predict the right value (\textit{1999}). 
	}
\end{figure}

\begin{figure*}[!t]
	\centering
	\fbox{\includegraphics[width=6in]{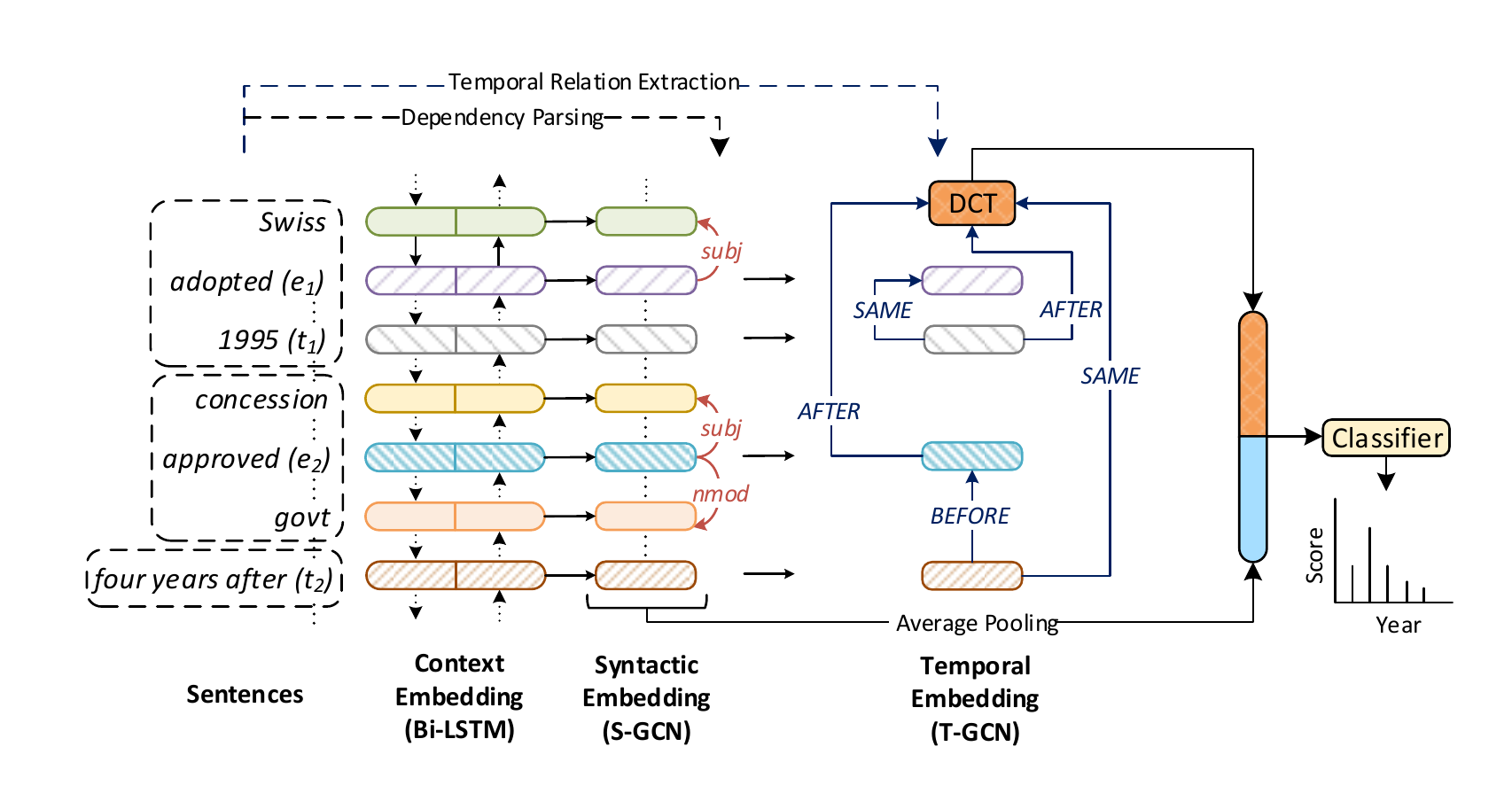}}
	\caption{\label{fig:overview_NeuralDater}Overview of \method{}. \method{} exploits syntactic and temporal structure in a document to learn effective representation, which in turn are used to predict the  document time. \method{} uses a Bi-directional LSTM (Bi-LSTM), two Graph Convolution Networks (GCN) -- one over the dependency tree and the other over the document's temporal graph -- along with a softmax classifier, all trained end-to-end jointly.}	
\end{figure*}

Initial attempts on automatic document dating started with generative models by \cite{de_jong05}. This model is
later improved by \cite{Kanhabua:2008:ITL:1429852.1429902} incorporating additional features such as POS tags, collocations, etc. \cite{Chambers:2012:LDT:2390524.2390539} shows significant improvement over these prior efforts through their discriminative models using hand crafted temporal features. \cite{Kotsakos:2014:BAD:2600428.2609495} propose a statistical approach for document dating exploiting term burstiness \cite{Lappas:2009:BSD:1557019.1557075}. 

Document dating is a challenging  problem which requires extensive reasoning over the temporal structure of the document. Let us motivate this through an example shown in \reffig{fig:motivation}. In the document, \textit{four years after} plays a crucial role in identifying the creation time of the document. The existing approaches give higher confidence for time-stamp immediate to the year mention \textit{1995}. \method{}, \methode{} (\methodes{}) and \methodc{} (\methodcs{}) exploits  the syntactic and temporal structure of the document to predict the right time-stamp (1999) for the document. With the exception of \cite{Chambers:2012:LDT:2390524.2390539}, all prior work on the document dating problem ignore such informative temporal structure within the document.

Research in document event extraction and ordering have made it possible to extract such temporal structures involving events, temporal expressions, and the (unknown) document date in a document \cite{catena_paper,Chambers14}. While methods to perform reasoning over such structures exist  \cite{tempeval07,tempeval10,tempeval13,tempeval15,timebank03}, none of them have exploited   advances in deep learning \cite{alexnet,microsoft_speech,deep_learning_book}. In particular, recently proposed Graph Convolution Networks (GCN) \cite{Defferrard:2016:CNN:3157382.3157527,kipf2016semi} have emerged as a way to learn graph representation while encoding structural information and constraints represented by the graph. We adapt GCNs for the document dating problem.

Motivated by the effectiveness of attention based models in different NLP tasks \cite{attention_qna,bahdanau_attn}, we incorporated attention in our method in a principled fashion. We not only did some focused context capturing using attention but also incorporated it for feature aggregation \cite{graphsage} in the graph convolution network. 

 Our contributions for this task are many-fold:

\begin{itemize}
	\item We propose \emph{\method{}}, a Graph Convolution Network (GCN)-based approach for document dating. To the best of our knowledge, this is the first application of GCNs, and more broadly deep neural network-based methods, for the document dating problem.
	\item We propose  \emph{\methodfin{}}: Attentive Deep Document Dater, the first attention-based model for time-stamping documents.
	\item We devise a novel method for label based attentive graph convolution over directed graphs and use it for this task.
	\item Through extensive experiments on multiple real-world datasets, we demonstrate the effectiveness of our models over the previous baseline methods.
\end{itemize}
\subsection{\textbf{Related Work}}
\label{sec:related_work_NeuralDater}

{\bf Automatic Document Dating}:
\cite{de_jong05} propose the first approach for automating document dating through a statistical language model. \cite{Kanhabua:2008:ITL:1429852.1429902} further extend this work by incorporating semantic-based preprocessing and temporal entropy \cite{temporal_entropy} based term-weighting. 
\cite{Chambers:2012:LDT:2390524.2390539} propose a MaxEnt based discriminative model trained on hand-crafted temporal features. They also propose a model to learn probabilistic constraints between year mentions and the actual creation time of the document. We draw inspiration from their work for exploiting temporal reasoning for document dating. 
\cite{Kotsakos:2014:BAD:2600428.2609495} propose a purely statistical method which considers lexical similarity alongside burstiness \cite{Lappas:2009:BSD:1557019.1557075} of terms for dating documents. To the best of our knowledge, \method{} is the first method to utilize deep learning techniques and \methodfin{} is the first method to utilize attention based deep learning techniques for the document dating problem.



{\bf Event Ordering Systems}: Temporal ordering of events is a vast research topic in NLP. The problem is posed as a temporal relation classification between two given temporal entities. Machine Learned classifiers and well crafted linguitsic features for this task are used in \cite{Chambers:2007:CTR:1557769.1557820, E14-1033}. \cite{N13-1112} use a hybrid approach by adding 437 hand crafted rules. \cite{Chambers:2008:JCI:1613715.1613803, P09-1046} try to classify with many more temporal constraints, while utilizing integer linear programming and Markov logic. 

CAEVO, a CAscading EVent Ordering architecture \cite{Chambers14} use sieve-based architecture 
\cite{sieve_architecture} 
for temporal event ordering for the first time. They mix multiple learners according to their precision based ranks and use transitive closure for maintaining consistency of temporal graph. \cite{catena_paper} recently propose CATENA (CAusal and TEmporal relation extraction from NAtural language texts), the first integrated system for the temporal and causal relations extraction between pre-annotated events and time expressions. They also incorporate sieve-based architecture which outperforms existing methods in temporal relation classification domain. We make use of CATENA for temporal graph construction in our work. 

{\bf Graph Convolutional Networks (GCN)}:
GCNs generalize Convolutional Neural Network (CNN) over graphs. GCN is introduced by \cite{gcn_first_work}, and later extended by \cite{Defferrard:2016:CNN:3157382.3157527} with efficient localized filter approximation in spectral domain. \cite{kipf2016semi} propose a first-order approximation of localized filters through layerwise propagation rule. GCNs over syntactic dependency trees have been recently exploited in the field of semantic-role labeling \cite{gcn_srl}, neural machine translation \cite{gcn_nmt}, event detection \cite{gcn_event}. In our work, we successfully use GCNs for document dating.

{\bf Attention Network:}
Attention networks have been well exploited for various tasks like document classification \cite{hierarchical_attn}, question answering \cite{attention_qna}, machine translation \cite{bahdanau_attn, NIPS2017_7181} to name a few. Recently, attention over graph structure has been proved to work well as shown by \cite{graphAttention}. Taking motivation from them, we deploy an attentive graph convolutional network on temporal graph.

\subsection{\textbf{Background: Graph Convolution Networks (GCN)}}
\label{sec:background_NeuralDater}

In this section, we provide an overview of Graph Convolution Networks (GCN) \cite{kipf2016semi}. GCN learns an embedding for each node of the graph it is applied over. We first present GCN for undirected graphs and then move on to GCN for directed graph setting. 

\subsubsection{\textbf{GCN on Undirected Graph}}
\label{sec:undirected_gcn_NeuralDater}

Let $\m{G} = (\m{V}, \m{E})$ be an undirected graph, where $\m{V}$ is a set of $n$ vertices and $\m{E}$ the set of edges. The input feature matrix  $\m{X} \in \mathbb{R}^{n \times m}$ whose rows are input representation of node $u$, $x_{u} \in \mathbb{R}^{m}\text{, }\forall u \in \m{V}$. The output hidden representation $h_v \in \mathbb{R}^{d}$ of a node $v$ after a single layer of graph convolution operation can be obtained by considering only the immediate neighbours of  $v$. This can be formulated as:
$$h_{v} = f\left(\sum_{u \in \m{N}(v)}\left(W x_{u} + b\right)\right),~~~\forall v \in \m{V} .$$
Here, model parameters $W \in \mathbb{R}^{d \times m}$ and $b \in \mathbb{R}^{d}$ are learned in a task-specific setting using first-order gradient optimization. $\m{N}(v)$ refers to the set of neighbours of $v$ and $f$ is any non-linear activation function. We have used ReLU as the activation function in this paper\footnote{ReLU: $f(x) = \max(0, x)$}.  

In order to capture nodes many hops away, multiple GCN layers may be stacked one on top of another. In particular, $h^{k+1}_{v}$, representation of node $v$  after $k^{th}$ GCN layer can be formulated as:
$$h^{k+1}_{v} = f\left(\sum_{u \in \m{N}(v)}\left(W^{k} h^{k}_{u} + b^{k}\right)\right), \forall v \in \m{V} . $$
where $h^{k}_{u}$ is the input to the $k^{th}$ layer.\\

\subsubsection{\textbf{GCN on Labeled and Directed Graph}}
\label{sec:directed_gcn_NeuralDater}

In this section, we consider GCN formulation over graphs where each edge is labeled as well as directed. In this setting, an edge from node $u$ to $v$ with label $l(u, v)$ is denoted as $(u, v, l(u, v))$. While a few recent works focus on GCN over directed graphs \cite{gcn_summ,gcn_srl}, none of them consider labeled edges. We handle both direction and label by incorporating label and direction specific filters.

Based on the assumption that the information in a directed edge need not only propagate along its direction, following \cite{gcn_srl} we define an updated edge set $\m{E'}$ which expands the original set $\m{E}$ by incorporating inverse, as well self-loop edges.
\begin{multline}	
\m{E'} = \m{E} \cup \{(v,u,l(u,v)^{-1})~|~ (u,v,l(u,v)) \in \m{E}\}  \\
\cup \{(u, u, \top)~|~u \in \m{V})\} \label{eqn:updated_edges_NeuralDater}.
\end{multline}

Here, $l(u,v)^{-1}$ is the inverse edge label corresponding to label $l(u,v)$, and $\top$ is a special empty relation symbol for self-loop edges. We now define $h_{v}^{k+1}$ as the embedding of node $v$ after $k^{th}$ GCN layer applied over the directed and labeled graph as:
\begin{equation}
h_{v}^{k+1} = f \left(\sum_{u \in \m{N}(v)}\left(W^{k}_{l(u,v)}h_{u}^{k} + b^{k}_{l(u,v)}\right)\right).
\label{eqn:gcn_diretected_NeuralDater}
\end{equation}

We note that the parameters $W^{k}_{l(u,v)}$ and $b^{k}_{l(u,v)}$ in this case are edge label specific.

\subsubsection{\textbf{Incorporating Edge Importance}}
\label{sec:gating_NeuralDater}

In many practical settings, we may not want to give equal importance to all the edges. For example, in case of automatically constructed graphs, some of the edges may be erroneous and we may want to automatically learn to discard them. Label based edge-wise gating may be used in a GCN to give importance to relevant edges and subdue the noisy ones. \cite{gcn_event, gcn_srl} used gating for similar reasons and obtained high performance gain. At $k^{th}$ layer, we compute gating value for a particular labeled edge $(u,v, l(u,v))$ as:
\[
g^{k}_{l(u,v)} = \sigma \left( h^{k}_u \cdot \hat{w}^{k}_{l(u,v)} + \hat{b}^{k}_{l(u,v)} \right),
\]
where, $\sigma(\cdot)$ is the sigmoid function, $\hat{w}^{k}_{l(u,v)}$ and $ \hat{b}^{k}_{l(u,v)}$ are label specific gating parameters. Thus, gating helps to make the model robust to the noisy labels and directions of the input graphs. GCN embedding of a node while incorporating edge gating may be computed as follows.
\[
h^{k+1}_{v} = f\left(\sum_{u \in \m{N}(v)} g^{k}_{l(u,v)} \times \left({W}^{k}_{l(u,v)} h^{k}_{u} + b^{k}_{l(u,v)}\right)\right).
\]

\subsection{\textbf{\method{} Details}}
\label{sec:method_NeuralDater}

The Documents Dating problem may be cast as a multi-class classification problem \cite{Kotsakos:2014:BAD:2600428.2609495,Chambers:2012:LDT:2390524.2390539}. In this section, we present a detailed discussion of \method{}. Architectural overview of \method{} is shown in \reffig{fig:overview_NeuralDater}.

\method{} is a deep learning-based multi-class classification system. It takes in a document as input and returns its predicted date as output by exploiting the syntactic and temporal structure of document. 

\method{} network consists of three layers which learn an  embedding for the Document Creation Time (DCT) node corresponding to the document. This embedding is then fed to a softmax classifier which produces a distribution over timestamps. Following prior research \cite{Chambers:2012:LDT:2390524.2390539,Kotsakos:2014:BAD:2600428.2609495}, we work with year granularity for the experiments in this paper.  We, however, note that NeuralDater can be trained for finer granularity with appropriate training data. The \method{} network is trained end-to-end using training data. Each component is described in greater detail below:

\subsubsection{\textbf{Context Embedding (Bi-LSTM)}}
\label{sec:et_gcn}

Let us consider a document $D$ with $n$ tokens  $w_1, w_2, ..., w_n$.
We first represent each token by a $k$-dimensional word embedding. For the experiments in this paper, we use GloVe \cite{glove} embeddings. These token embeddings are stacked together to get the document representation $\m{X} \in \mathbb{R}^{n \times k}$. We then employ a Bi-directional LSTM (Bi-LSTM) \cite{lstm_1997} on the input matrix $\m{X}$ to obtain contextual embedding for each token. After stacking contextual embedding of all these tokens, we get the new document representation matrix $\m{H}^{cntx} \in \mathbb{R}^{n \times r_{cntx}}$. In this new representation, each token is represented in a $r_{cntx}$-dimensional space. Our choice of LSTMs for learning contextual embeddings for tokens is motivated by the previous success of LSTMs in this task \cite{Sutskever:2014:SSL:2969033.2969173}.

\subsubsection{\textbf{Syntactic Embedding (S-GCN)} }
\label{sec:syntax_gcn_NeuralDater}

While the Bi-LSTM is effective at capturing immediate local context of a token, it may not be as effective in capturing longer range dependencies among words in a sentence. For example, in \reffig{fig:motivation}, we would like the embedding of token ``approved'' to be directly affected by ``govt'', even though they are not immediate neighbours. A dependency parse may be used to capture such longer-range connections. In fact, similar features were exploited by \cite{Chambers:2012:LDT:2390524.2390539} for the document dating problem. \method{} captures such longer-range information by using a GCN run over the syntactic structure of the document. We describe this in detail below.

The context embedding, $\m{H}^{cntx} \in \mathbb{R}^{n \times r_{cntx}}$ learned in the previous step is used as input to this layer. For a given document, we first extract its syntactic dependency structure by applying the Stanford CoreNLP's dependency parser \cite{stanford_corenlp} on each sentence in the document individually. We now employ the Graph Convolution Network (GCN) over this dependency graph using the GCN formulation presented in \refsec{sec:directed_gcn_NeuralDater}. We call this GCN the Syntactic GCN or S-GCN.

Since S-GCN operates over the dependency graph and uses \refeqn{eqn:gcn_diretected_NeuralDater} for updating embeddings, the number of parameters in S-GCN is directly proportional to the number of dependency edge types. Stanford CoreNLP's dependency parser returns 55 different dependency edge types. This large number of edge types is going to significantly over-parameterize S-GCN, thereby increasing the possibility of overfitting. In order to address this, we use only three edge types in S-GCN. For each edge connecting nodes $w_i$ and $w_j$ in $\m{E'}$ (see \refeqn{eqn:updated_edges_NeuralDater}), we determine its new type $L(w_i, w_j)$ as follows:
\begin{itemize}
	\item $L(w_i, w_j) =\ \rightarrow$ if $(w_i, w_j, l(w_i, w_j)) \in \m{E'}$, i.e., if the edge is an original dependency parse edge
	\item $L(w_i, w_j) =\ \leftarrow$ if $(w_i, w_j, l(w_i, w_j)^{-1}) \in \m{E'}$, i.e., if the edges is an inverse edge	
	\item $L(w_i, w_j) =\ \top$ if $(w_i, w_j, \top) \in \m{E'}$, i.e., if the edge is a self-loop with $w_i = w_j$
\end{itemize}
S-GCN now estimates embedding $h^{syn}_{w_{i}} \in \mathbb{R}^{r_{syn}}$ for each token $w_{i}$ in the document using the formulation shown below.
\[
\Scale[0.93]{h^{syn}_{w_i} = f \Bigg(\sum_{w_j \in \m{N}(w_i)}g_{L(w_i, w_j)} \left(W_{L(w_i, w_j)}h^{cntx}_{w_j} + b_{L(w_i, w_j)}\right) \Bigg)}
\]
Please note S-GCN's use of the new edge types $L(w_i, w_j)$  above, instead of the $l(w_i, w_j)$ types used in \refeqn{eqn:gcn_diretected_NeuralDater}. By stacking embeddings for all the tokens together, we get the new embedding matrix $\m{H}^{syn} \in \mathbb{R}^{n \times r_{syn}}$ representing the document.

\textbf{AveragePooling}: We obtain an embedding $h_{D}^{avg}$ for the whole document by average pooling of every token representation.
\begin{equation}
h_{D}^{avg} = \frac{1}{n} \sum_{i = 1}^{n} h_{w_i}^{syn}
\label{eqn:avg-pool}.
\end{equation}


\subsubsection{\textbf{Temporal Embedding (T-GCN)}}
\label{sec:t-gcn_NeuralDater}

In this layer, \method{} exploits temporal structure of the document to learn an embedding for the Document Creation Time (DCT) node of the document. First, we describe the construction of temporal graph, followed by GCN-based embedding learning over this graph.

\textbf{Temporal Graph Construction}: \method{} uses Stanford's SUTime tagger \cite{sutime} for date normalization and the event extraction classifier of \cite{Chambers14} for event detection. The annotated document is then passed to CATENA \cite{catena_paper}, current state-of-the-art temporal and causal relation extraction algorithm, to obtain a temporal graph for each document. 
Since our task is to predict the creation time of a given document, we supply DCT as unknown to CATENA. We hypothesize that the temporal relations extracted in absence of DCT are helpful for document dating and we indeed find this to be true, as shown in Section \ref{sec:results_AD3}. Temporal graph is a directed graph, where nodes correspond to events, time mentions, and the Document Creation Time (DCT). Edges in this graph represent causal and temporal relationships between them. Each edge is attributed with a label representing the type of the temporal relation. CATENA outputs 9 different types of temporal relations, out of which we selected five types, viz.,  \textit{AFTER}, \textit{BEFORE}, \textit{SAME}, \textit{INCLUDES}, and  \textit{IS\_INCLUDED}. The remaining four types were ignored as they were substantially infrequent. 

Please note that the temporal graph may involve only a small number of tokens in the document. For example, in the temporal graph in \reffig{fig:overview_NeuralDater}, there are a total of 5 nodes: two temporal expression nodes (\textit{1995} and \textit{four years after}), two event nodes (\textit{adopted} and \textit{approved}), and a special DCT node. This graph also consists of temporal relation edges such as (\textit{SAME}, \textit{AFTER}, \textit{BEFORE}). 


\textbf{Temporal Graph Convolution}: \method{} employs a GCN over the temporal graph constructed above. We refer to this GCN as the Temporal GCN or T-GCN. T-GCN is based on the GCN formulation presented in \refsec{sec:directed_gcn_NeuralDater}. Unlike S-GCN, here we consider label and direction specific parameters as the temporal graph consists of only five types of edges.

Let $n_T$ be the number of nodes in the temporal graph. Starting with $\m{H}^{syn}$ (\refsec{sec:syntax_gcn_NeuralDater}), T-GCN learns a $r_{temp}$-dimensional embedding for each node in the temporal graph. Stacking all these embeddings together, we get the embedding matrix $\m{H}^{temp} \in \mathbb{R}^{n_{T} \times r_{temp}}$. T-GCN embeds the temporal constraints induced by the temporal graph in $h_{DCT}^{temp} \in \mathbb{R}^{r_{temp}}$, embedding of the DCT node of the document. 


\subsubsection{\textbf{Classifier}}
Finally, the DCT embedding $h_{DCT}^{temp}$ and average-pooled syntactic representation $h_{D}^{avg}$ (see \refeqn{eqn:avg-pool}) of document $D$ are concatenated and fed to a fully connected feed forward network followed by a softmax. This allows the \method{} to exploit context, syntactic, and temporal structure of the document to  predict the final document date $y$.
\begin{eqnarray*}
	h_{D}^{avg+temp} &=& \text{ } [h_{DCT}^{temp}~;~h_{D}^{avg}] \\ 
	p(y \vert D) &=& \mathrm{Softmax}(W \cdot h_{D}^{avg+temp} + b).
\end{eqnarray*}

Even though the previous discussion is presented in a sequential manner, the whole network is trained in a joint end-to-end manner using back propagation.

\subsection{\textbf{Attentive Deep Document Dater (\methodfin{}): Proposed Method}}
\label{sec:model_AD3}

\begin{figure*}[!t]
	\centering
	\fbox{\includegraphics[width=7in]{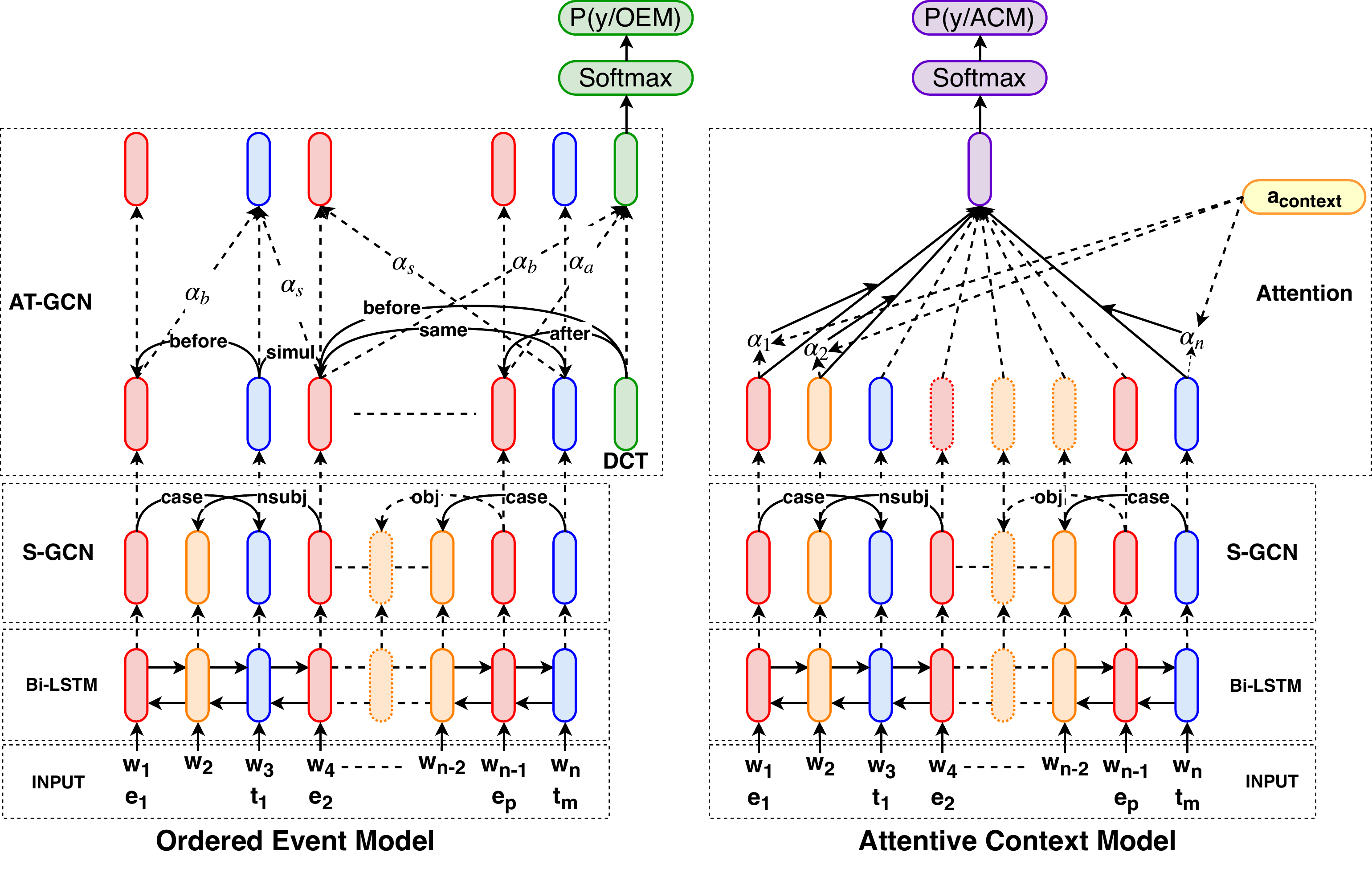}}
	\caption{\label{fig:overview_AD3} Two proposed models a) \methode{} (left) and b) \methodc{} (right), where $\text{w}_{\text{i}}$ are the words of a document (D), $\text{e}_{\text{i}}$ are the words signifying events and $\text{t}_{\text{i}}$ are the temporal tokens as detected by CATENA. Both the models use Bi-LSTM and S-GCN (Syntactic-GCN) in the initial part of their pipeline. \methode{} (\methodes{}) uses a label based attentive graph convolutional network for encoding the DCT, whereas \methodc{} (\methodcs{}) uses a word attention based model to encode the document. $\alpha_{i} \ \forall \ i \in [1,n]$ denotes the attentions over the words of document and $\alpha_{a}, \ \alpha_{b} $ and $\alpha_{s}$ denotes the attention over nodes connected with edge labels \textit{AFTER}, \textit{BEFORE} and \textit{SAME} respectively. \methodes{} provides the probability scores over the years given the encoded DCT, while \methodcs{} provides the probability scores given the context of the document. Both the models are trained separately.}
	
\end{figure*}
In this section, we describe Attentive Deep Document Dater (\methodfin{}). \methodfin{} is inspired by \method{}, and shares many of its components. Just like in \method{}, \methodfin{} also leverages two main types of signals from the document -- syntactic and event-time --  to predict the document's timestamp. However, there are crucial differences between the two systems. Firstly, instead of concatenating embeddings learned from these two sources, \methodfin{} treats these two models completely separate, and combines them at a later stage. Secondly, unlike NeuralDater, \methodfin{} employs attention mechanisms in each of these two models. We call the resulting models \methodc{} (\methodcs) and \methode{} (\methodes{}). These two models are described in \refsec{sec:reader_AD3} and \refsec{sec:reasoner_AD3}, respectively.

\subsubsection{\textbf{\methodc{} (\methodcs)}}
\label{sec:reader_AD3}

Recent success of attention-based deep learning models for classification \cite{hierarchical_attn}, question answering \cite{attention_qna}, and  machine translation \cite{bahdanau_attn}  have motivated us to use attention during document dating. We extend the syntactic embedding model of NeuralDater (\refsec{sec:syntax_gcn_NeuralDater}) by incorporating an attentive pooling layer. 
We call the resulting model \methodcs{}. This model (right side in \reffig{fig:overview_AD3}) has two major components.
\begin{itemize}
	\item\textbf{Context Embedding and Syntactic Embedding}:  Following NeuralDater, we used Bi-LSTM and S-GCN to capture context and long range syntactic dependencies in the document. The syntactic embedding, $H^{syn} \in \mathbb{R}^{n \times k_{syn}}$ is then fed to an  Attention Network for further processing. Note that, $k_{syn}$ is the dimension of the output of Syntactic-GCN and $n$ is the number of tokens in the document.
	
	\item \textbf{Attentive Embedding}: In this layer we learn the representation for the whole document through word level attention network. We learn a context vector, $u_{s} \in \mathbb{R}^{s}$ with respect to which we calculate attention for each token. Finally, we aggregate the token features with respect to their attention weights in order to represent the document. More formally, let $h^{syn}_{t} \in \mathbb{R}^{syn}$ be the  syntactic representation of the $t^{th}$ token in the document. We take non-linear projection of it in $\mathbb{R}^{s}$ with $W_{s} \in \mathbb{R}^{syn \times s}$. Attention weight $\alpha_{t}$ for  $t^{th}$ token is calculated with respect to the context vector $u_{t}^{T}$ as follows:
	$$u_{t} = \mathrm{tanh}(W_{s}h^{syn}_{t}),$$
	$$\alpha_{t} = \frac{\mathrm{exp}(u_{t}^{T}u_{s})}{\sum_{t}\mathrm{exp}(u_{t}^{T}u_{s})}.$$
	
	Finally, the document representation for the \methodcs{} is computed as shown below.
	\[
	d_{\mathrm{\methodcs{}}} = \sum_{t}\alpha_{t}h^{syn}_{t}
	\]
	This representation is fed to a fully connected feed forward network followed by a softmax. 
\end{itemize}
The final probability distribution over years predicted by the \methodcs{} is given below.
$$\mathrm{P}_{\mathrm{\methodcs{}}}(y \vert D) = \mathrm{Softmax}(W \cdot d_{\mathrm{\methodcs{}}} + b).$$
%

\subsubsection{\textbf{\methode{} (\methodes{})}}
\label{sec:reasoner_AD3}

The \methodes{} model is shown on the left side of \reffig{fig:overview_AD3}. Just like in \methodcs{}, context and syntactic embedding is also part of \methodes{}. The syntactic embedding is fed to the Attentive Graph Convolution Network (AT-GCN) where the  graph is obtained from the event-time ordering algorithm CATENA \cite{catena_paper}. We describe these components in detail below.

\subsubsection{\textbf{Temporal Graph }}

We use the same process used in NeuralDater \cite{NeuralDater} for procuring the Temporal Graph from the document. 
Let $\m{E}_{T}$ be the edge list of the Temporal Graph. Similar to \cite{gcn_srl,NeuralDater}, we add reverse edges for each of the existing edge and self loops for passing current node information as explained in \refsec{sec:directed_gcn_NeuralDater}. The new edge list $\m{E}_{T}^{'}$ is shown below.
\begin{multline*}    
\m{E'_{\text{T}}} = \m{E_{\text{T}}} \cup \{(j,i,l(i,j)^{-1})~|~ (i,j,l(i,j)) \in \m{E_{\text{T}}}\} \\
\cup \{(i, i, \mathrm{self})~|~i \in \m{V})\}.
\end{multline*}
%

Let $\m{L}$ be the set of labels after the extra edges were added.
\subsubsection {\textbf{Attentive Graph Convolution (AT-GCN)} }
\label{sec:graph_atten}

Since the temporal graph is automatically generated, it is likely to have incorrect edges. Ideally, we would like to minimize the influence of such noisy edges while computing temporal embedding. In order to suppress the noisy edges in the Temporal Graph and detect important edges for reasoning, we use attentive graph convolution \cite{graphsage} over the Event-Time graph. We note that, gating, as used in \method{} is label based and does not take the neighbouring nodes into account for judging the importance of the edges. The attention mechanism learns the aggregation function jointly during training. Here, the main objective is to calculate the attention over the neighbouring nodes with respect to the current node for a given label. Then the embedding of the current node is updated by mixing neighbouring node embedding according to their attention scores. In this respect, we propose a label-specific attentive graph convolution over directed graphs.

Let us consider an edge in the temporal graph from node $i$ to node $j$ with type $l$, where $l \in \m{L}$ and $\m{L}$ is the label set. The  label set $\m{L}$ can be divided broadly into three coarse labels as done in \refsec{sec:syntax_gcn_NeuralDater}, Out of which the attention weights are specific to only two type of edges to reduce parameter and prevent overfitting, leaving out the self-loop edges. 
For illustration, if there exists an edge from node $i$  to $j$ then the edge types will be,
\begin{itemize}
	\item $ L(i, j) =\  \rightarrow$ if $(i, j, l(i, j)) \in \m{E}_{T}^{'}$,\\ i.e., if the edge is an original event-time edge.
	\item  $L(i, j) =\  \leftarrow$ if $(i, j, l(i, j)^{-1}) \in \m{E}_{T}^{'}$,\\ i.e., if the edge is added later.
\end{itemize}
First, we take a linear projection ($W^{atten}_{L(i,j)} \in \mathbb{R}^{k_{syn} \times F}$) of both the nodes in order to map both of them in the same direction-specific space. The concatenated vector $ [W_{L(i,j)}^{atten}\times h_{i}; W_{L(i,j)}^{atten} \times h_{j} ]$, signifies the importance of the node $j$ w.r.t node $i$. A non linear transformation of this concatenation can be treated as the importance feature vector between $i$ and $j$, \\
$$ e_{ij} = \mathrm{tanh} [W^{atten}_{L(i,j)} \times h_{i}; W^{atten}_{L(i,j)} \times h_{j} ].$$
\begin{figure}[t]
	\centering
	\includegraphics[width=3in]{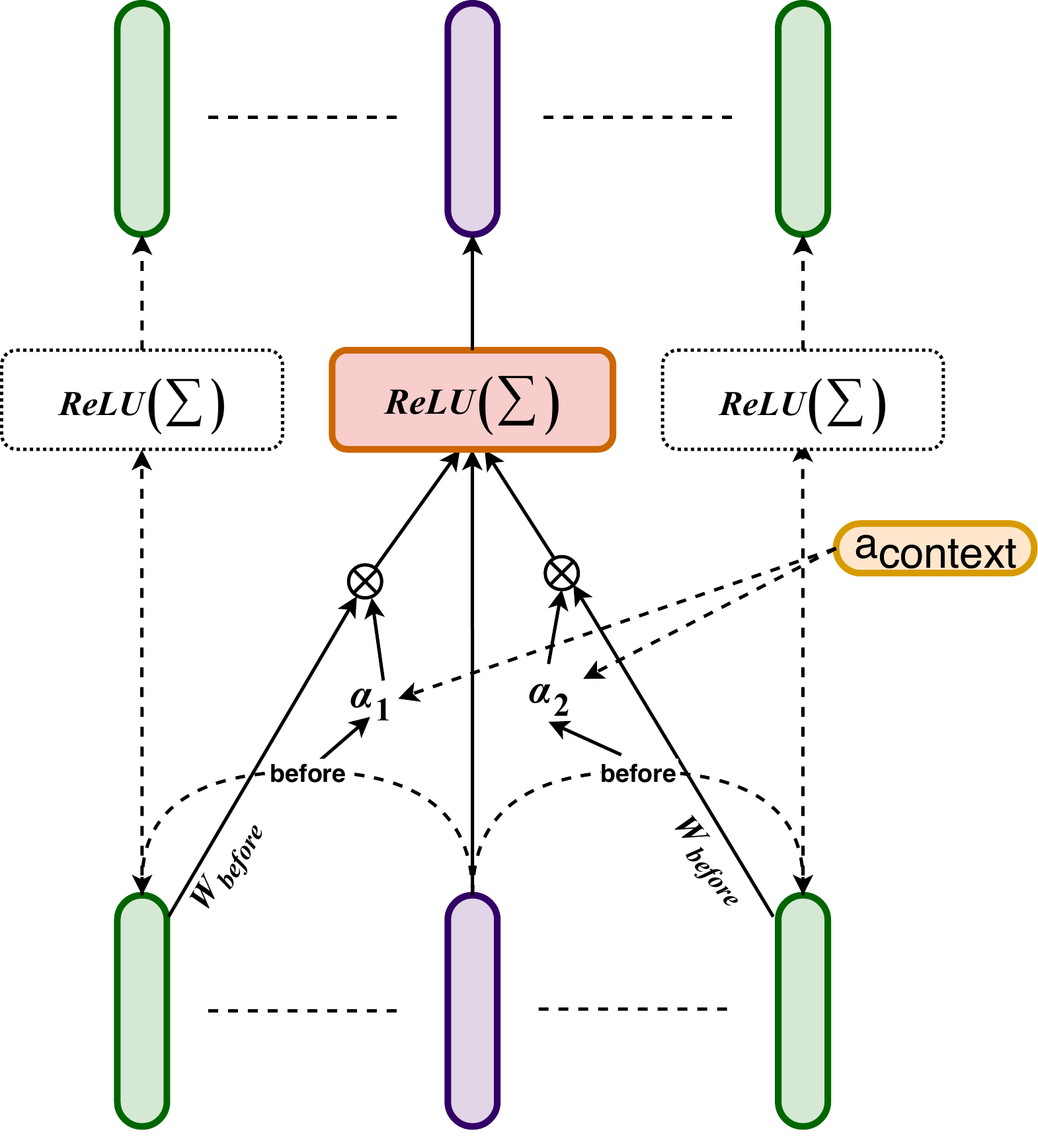}
	\caption{\label{fig:Attentive_Graph_Convolution} Attentive Graph Convolution (AT-GCN). In this layer, we learn attention weights for every edge based on label and direction. The attention weights are learnt using a context vector. The final representation of every node is a summation of weighted convolution over neighbouring nodes based on labels.
	}
\end{figure}
Now, we compute the attention weight of node $j$ for node $i$  with respect to a direction-specific context vector $a_{L(i,j)} \in \mathbb{R}^{2F}$, as follows: 
\[
\alpha^{l(i, j)}_{ij} = \frac{\text{exp}\left(a_{L(i,j)}^{T} e_{ij}   \right) }{\sum\limits_{  k \in \m{N}^{l(i,\cdot)}_{i}}\text{exp} \left(a_{L(i,j)}^{T} e_{ik}   \right)  },
\label{eqn:attention_cal}
\]
where, $\alpha^{l(i, j)}_{ij} = 0$ if node $i$ and $j$ is not connected through label $l$. $\m{N}^{l(i,\cdot)}$ denotes the subset of the neighbourhood of node $i$ with label $l$ only. Please note that, although the linear transform weight ($ W^{atten}_{L(i,j)}  \in \mathbb{R}^{k_{syn} \times F}$) is specific to the coarse labels $L$, but for each finer label $l \in \m{L}$ we get these convex weights of attentions. \reffig{fig:Attentive_Graph_Convolution} illustrates the above description w.r.t edge type \textit{BEFORE}.

Finally, the feature aggregation is done according to the attention weights. Prior to that, another label specific linear transformation is taken to perform the convolution operation. Then, the updated feature for node $i$ is carried out as follows.
\[
\Scale[1]{ h_{i}^{k+1} = f \left( \sum_{l \in \m{L}} \sum_{j \in \m{N}^{l(i,\cdot)}_{i}}\alpha^{l(i, j)}_{ij}\left(W_{l(i,j)}h_{j} + b_{l(i,j)}\right)\right).
	\label{eqn:graph_atten}}
\]
where, $ \alpha_{ii} = 1$. Note that,  $\alpha^{l(i, j)}_{ij} = 0$ when $j \notin \m{N}^{l(i,\cdot)}$. To illustrate formally, from \reffig{fig:Attentive_Graph_Convolution}, we see that weight $\alpha_{1}$ and $\alpha_{2}$ is calculated specific to label type \textit{BEFORE} and the neighbours which are connected through \textit{BEFORE} is being multiplied with $W_{before}$ prior to aggregation in the $ReLU$ block.

Now, after applying attentive graph convolution network, we only consider the representation of Document Creation Time (DCT),  $h_{DCT}$, as the document representation itself. $h_{DCT}$ is now passed through a fully connected layer prior to softmax. Prediction of the \methodes{} for the document D will be given as
$$\mathrm{P}_{\mathrm{\methodes}}(y \vert D) = \mathrm{Softmax}(W \cdot d_{\mathrm{DCT}} + b).$$

\subsubsection{\textbf{\methodfin{}: Attentive Deep Document Dater} }
\label{sec:read_reason}

\begin{figure}[t]
	\centering
	\includegraphics[width=3in]{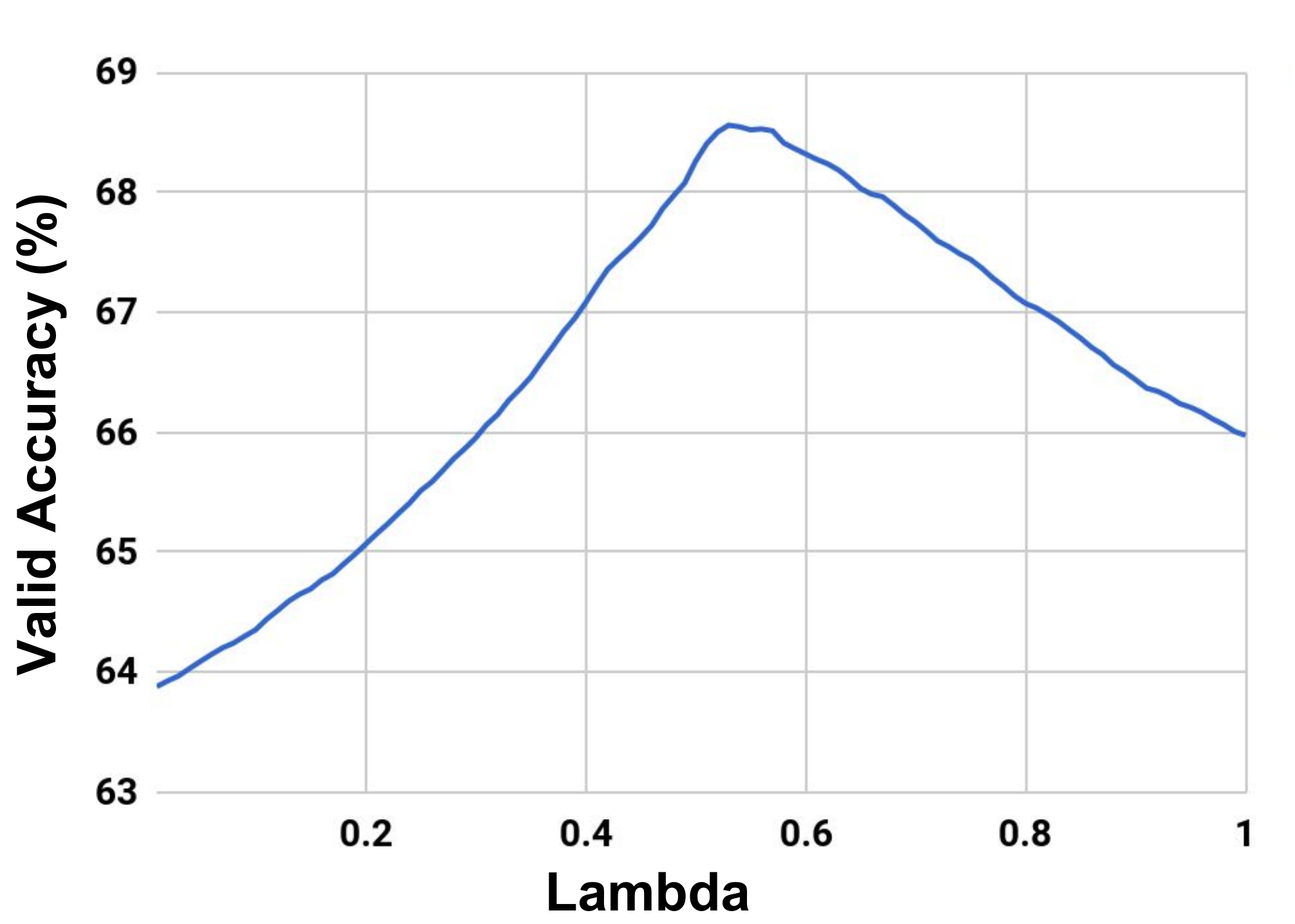}
	\caption{\label{fig:Accuracy_vs_Lambda} Variation of validation accuracy with $\lambda$ (for APW dataset). We observe that \methodcs{} and \methodes{} are both important for the task as we get optimal $\lambda$ = 0.52.
	}
\end{figure}

In this section, we propose an unified model by mixing both \methodcs{} and \methodes{}. The completely different learning paradigm of the two models motivated the unification of them. We take convex combination of the output probabilities of the models. 
\begin{multline*}
$$\mathrm{P}_{joint}(y \vert D) = \lambda\mathrm{P}_{\mathrm{\methodcs}}(y \vert D) \\
+(1-\lambda) \mathrm{P}_{\mathrm{\methodes}}(y \vert D)$$.
\end{multline*}

The combination hyper-parameter $\lambda$ is tuned on the validation data. 
We obtain the value of $\lambda$ to be 0.52 (\reffig{fig:Accuracy_vs_Lambda}) and 0.54 for APW and NYT datasets, respectively. This depicts that the two models are capturing significantly different aspects of documents and both the models are equally important for the task, resulting in a substantial improvement in performance when combined.

\subsection{\textbf{Experimental Setup}}
\label{sec:experiments_AD3}

\begin{table}[t]
	\centering
	\begin{tabular}{cccc}
		\toprule
		Datasets 	& \# Docs & Start Year & End Year\\
		\midrule
		APW 		&  675k	& 1995  & 2010 \\
		NYT			&  647k	& 1987  & 1996 \\
		\bottomrule
		\addlinespace
	\end{tabular}
	\caption{\label{tb:datasets_AD3}Details of datasets used. Please see \refsec{sec:experiments_AD3} for details.}
\end{table}


\textbf{Datasets}: We experiment on Associated Press Worldstream (APW) and New York Times (NYT) sections of Gigaword corpus \cite{gigaword5th}. The original dataset contains around 3 million documents of APW and 2 million documents of NYT spanning multiple years. From both sections, we randomly sample around 650k documents while maintaining balance among years. Documents belonging to years with substantially fewer documents are omitted. Details of the dataset can be found in Table \ref{tb:datasets_AD3}. For train, test and validation splits, the dataset was randomly divided in 8:1:1 ratio.

\textbf{Evaluation Criteria}: Given a document, the model needs to predict the year in which the document was published. We measure performance in terms of overall accuracy and mean absolute deviation of the model predictions from the true class.

\textbf{Baselines}: For evaluating our models, we compared against the following methods:

\begin{itemize}
	\item \textbf{BurstySimDater} \cite{Kotsakos:2014:BAD:2600428.2609495}:  This is a purely statistical method which uses lexical similarity and term burstiness \cite{Lappas:2009:BSD:1557019.1557075} for dating documents in arbitrary length time frame. For our experiments, we took the time frame length as 1 year. Please refer to \cite{Kotsakos:2014:BAD:2600428.2609495} for more details.
	\item \textbf{MaxEnt-Time-NER}: Maximum Entropy (MaxEnt) based classifier trained on hand-crafted temporal and Named Entity Recognizer (NER) based features. More details in \cite{Chambers:2012:LDT:2390524.2390539}. 
	\item \textbf{MaxEnt-Joint}: Refers to MaxEnt-Time-NER combined with year mention classifier as described in \cite{Chambers:2012:LDT:2390524.2390539}. 
	\item \textbf{MaxEnt-Uni-Time:} MaxEnt based discriminative model which takes bag-of-words representation of input document with normalized time expression as its features. 
	\item \textbf{CNN:} A Convolution Neural Network (CNN) \cite{cnn_paper} based text classification model proposed by \cite{yoon_kim}, which attained state-of-the-art results in several domains. 
\end{itemize}

\textbf{Hyperparameters}: By default, edge gating (\refsec{sec:gating_NeuralDater}) is used in all GCNs for \method{}. The parameter $K$ represents the number of layers in T-GCN (\refsec{sec:t-gcn_NeuralDater}). We use 300-dimensional GloVe embeddings and 128-dimensional hidden state for both GCNs and BiLSTM with $0.8$ dropout for all our models. We used Adam \cite{adam_optimizer} with $0.001$ learning rate for training. For \method{} we have used $K = 1$ and for \methodes{} we have used $K = 2$. All our models have one layer of S-GCN.

\begin{table}[!t]
	\begin{small}
		\centering
		\begin{tabular}{lc}
			\toprule
			Method 			 & Accuracy \\
			\midrule		
			\addlinespace
			T-GCN 								& 57.3 \\
			S-GCN + T-GCN $(K=1)$				& 57.8 \\
			S-GCN + T-GCN $(K=2)$				& 58.8 \\
			S-GCN + T-GCN $(K=3)$				& \textbf{59.1} \\
			\midrule
			Bi-LSTM 							& 58.6 \\
			Bi-LSTM + CNN 						& 59.0 \\
			Bi-LSTM + T-GCN						& 60.5 \\
			Bi-LSTM + S-GCN + T-GCN (no gate)	& 62.7 \\
			Bi-LSTM + S-GCN + T-GCN $(K=1)$		& \textbf{64.1}  \\
			Bi-LSTM + S-GCN + T-GCN $(K=2)$		&63.8 \\
			Bi-LSTM + S-GCN + T-GCN $(K=3)$		& 63.3 \\
			\bottomrule
		\end{tabular}
		\caption{\label{tb:result_ablation}Accuracies (\%) of different ablated methods on the APW dataset. Overall, we observe that incorporation of context (Bi-LSTM), syntactic structure (S-GCN) and temporal structure (T-GCN) in \method{} achieves the best performance. Please see \refsec{sec:perf_comp_AD3} for details.}
	\end{small}
\end{table}

\subsection{\textbf{Results}}
\label{sec:results_AD3}

\subsubsection{\textbf{Ablation Comparisons}}
\label{sec:ablation}

For demonstrating the efficacy of GCNs and BiLSTM for the problem, we evaluate different ablated variants of \method{} on the APW dataset. Specifically, we validate the importance of using syntactic and temporal GCNs and the effect of eliminating BiLSTM from the model. Overall results are summarized in Table \ref{tb:result_ablation}. The first block of rows in the table corresponds to the case when BiLSTM layer is excluded from \method{}, while the second block denotes the case when BiLSTM is included. We also experiment with multiple stacked layers of T-GCN (denoted by $K$) to observe its effect on the performance of the model. 

We observe that embeddings from Syntactic GCN (S-GCN) are much better than plain GloVe embeddings for T-GCN as S-GCN encodes the syntactic neighborhood information in event and time embeddings which makes them more relevant for document dating task.

Overall, we observe that including BiLSTM in the model improves performance significantly. Single BiLSTM model outperforms all the models listed in the first block of  \reftbl{tb:result_ablation}. Also, some gain in performance is observed on increasing the number of T-GCN layers ($K$) in absence of BiLSTM, although the same does not follow when BiLSTM is included in the model. This observation is consistent with \cite{gcn_srl}, as multiple GCN layers become redundant in the presence of BiLSTM. We also find that eliminating edge gating from our best model deteriorates its overall performance.

In summary, these results validate our hypothesis that joint incorporation of syntactic and temporal structure of a document in \method{} results in improved performance.

\subsubsection{\textbf{Performance Comparison}}
\label{sec:perf_comp_AD3}

\begin{figure}[t]
	\centering
	\includegraphics[width=3.4in]{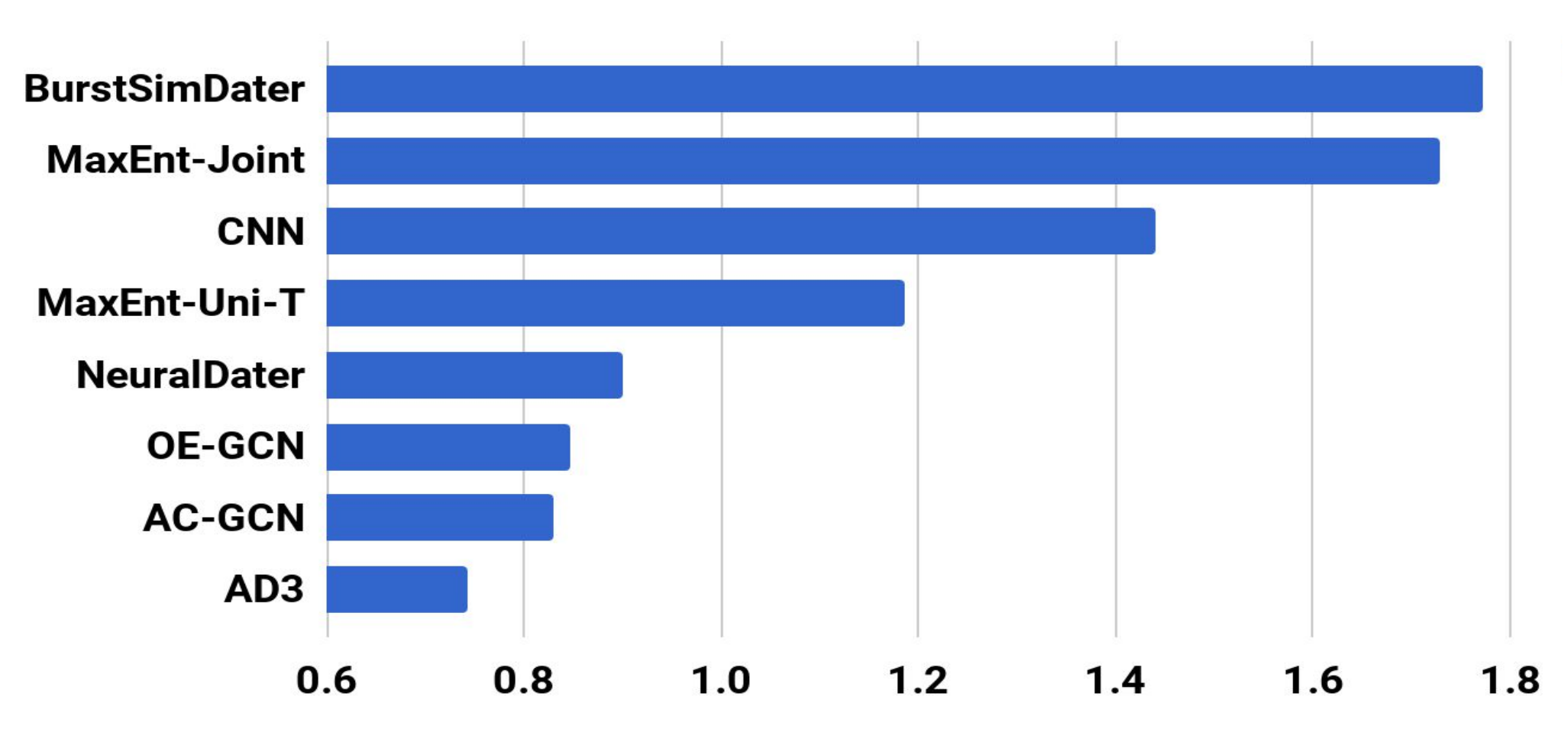}
	\caption{\label{fig:results_mean_dev}Mean absolute deviation (in years; lower is better) between a model's top prediction and the true year in the APW dataset. We find that all of our proposed methods outperform the previous models. Please refer to \refsec{sec:perf_comp_AD3} for details. 
	}
\end{figure}

\begin{table}[t]
	\centering
	\begin{tabular}{lcc}
		\toprule
		Method 			 & APW & NYT \\
		\midrule		
		\addlinespace
		BurstySimDater 		& 45.9 & 38.5 \\
		MaxEnt-Time+NER		& 52.5 & 42.3 \\
		MaxEnt-Joint		& 52.5 & 42.5 \\
		MaxEnt-Uni-Time		& 57.5 & 50.5 \\
		CNN 				& 56.3 & 50.4 \\
		\method			& 64.1 & 58.9 \\
		\hline
		\addlinespace
		\methodnew{} & 66.2 & 60.1\\
		\methodes{} [\ref{sec:reasoner_AD3}]& 63.9 & 58.3\\
		\methodcs{}  [\ref{sec:reader_AD3}]& 65.6 & 60.3\\
		\hline
		\addlinespace
		\textbf{\methodfin{}} [\ref{sec:read_reason}] & \textbf{68.2} & \textbf{62.2} \\
		\bottomrule
	\end{tabular}
	\caption{\label{tb:result_main}Accuracy (\%) of different methods on the APW and NYT datasets for the document dating problem (higher is better). The unified model significantly outperforms all previous models.}
\end{table}
In order to evaluate the effectiveness of our proposed methods, we compare it against existing document dating systems and text classification models. The final results are summarized in Table \ref{tb:result_main}. Overall, we find that our methods outperform all other methods with a significant margin on both datasets. We observe only a slight gain in the performance of MaxEnt-based model (MaxEnt-Time+NER) of \cite{Chambers:2012:LDT:2390524.2390539} on combining with temporal constraint reasoner (MaxEnt-Joint). This may be attributed to the fact that the model utilizes only year mentions in the document, thus ignoring other relevant signals which might be relevant to the task. BurstySimDater performs considerably better in terms of precision compared to the other baselines,  although it significantly underperforms in accuracy. We note that \methodfin{} outperforms all these prior models both in terms of accuracy and mean absolute deviation. We find that even generic deep-learning based text classification models, such as CNN \cite{yoon_kim}, are quite effective for the problem. However, since such a model doesn't give specific attention to temporal features in the document, its performance remains limited. From \reffig{fig:results_mean_dev}, we observe that \methodfin{}'s top prediction achieves on average the lowest deviation from the true year.

Among individual models \methodes{} performs at par with \method{}, while \methodcs{} outperforms it. The empirical results imply that \methodcs{} by itself is effective for this task. The relatively worse performance of \methodes{} can be attributed to the fact that it only focuses on the Event-Time information and leaves out most of the contextual information. However, both captures various different aspects of the document for classification, which motivated us to propose an ensemble of the two models. This explains the significant boost in performance of \methodfin{} over NeuralDater as well as the individual models. It is worth mentioning that although \methodcs{} and \methodes{} do not provide significant boosts in accuracy, their predictions have considerably lower mean-absolute-deviation as shown in \reffig{fig:results_mean_dev}.

We concatenated the DCT embedding provided by \methodes{} with the document embedding provided by \methodcs{} and trained in an end to end joint fashion like \method{}. We see that even with similar training method the \methodnew{} model on an average, performs 1.6\% better in terms of accuracy, once again proving the efficacy of attention based models over normal models.

\subsubsection{\textbf{Effectiveness of Attention}}
\label{sec:atten_perf}

\begin{figure}[t!]
	\centering
	\includegraphics[width=3in]{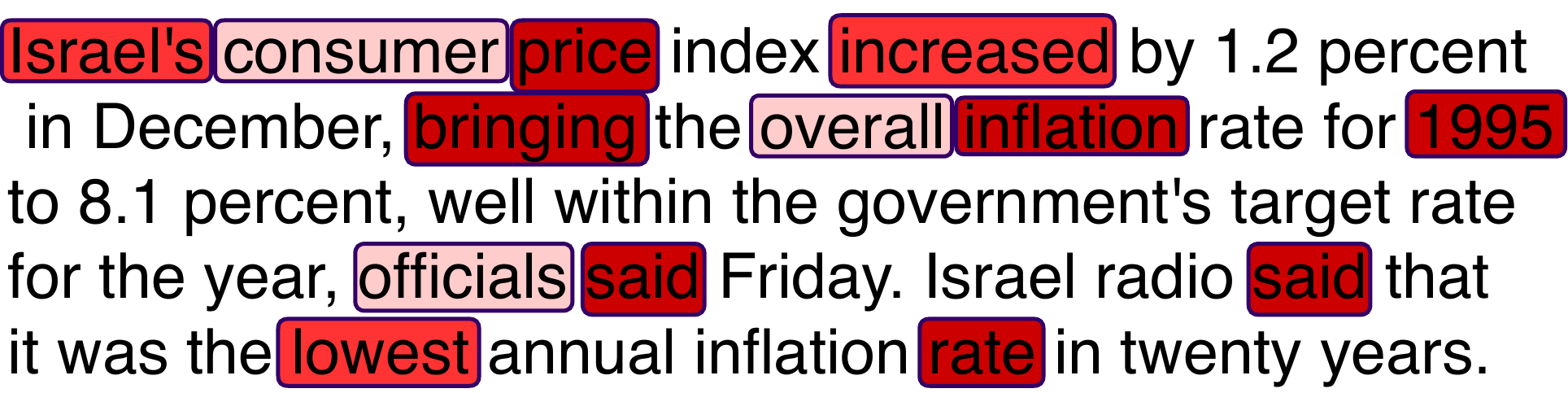}
	\caption{\label{fig:Context_Attention} Visualization of the attention of \methodcs{}. \methodcs{} captures the intuitive tokens as seen in the figure. Brighter colour implies higher attention. The correct DCT is 1996.
	}
\end{figure}
\begin{figure}[t!]
	\centering
	\includegraphics[width=3in]{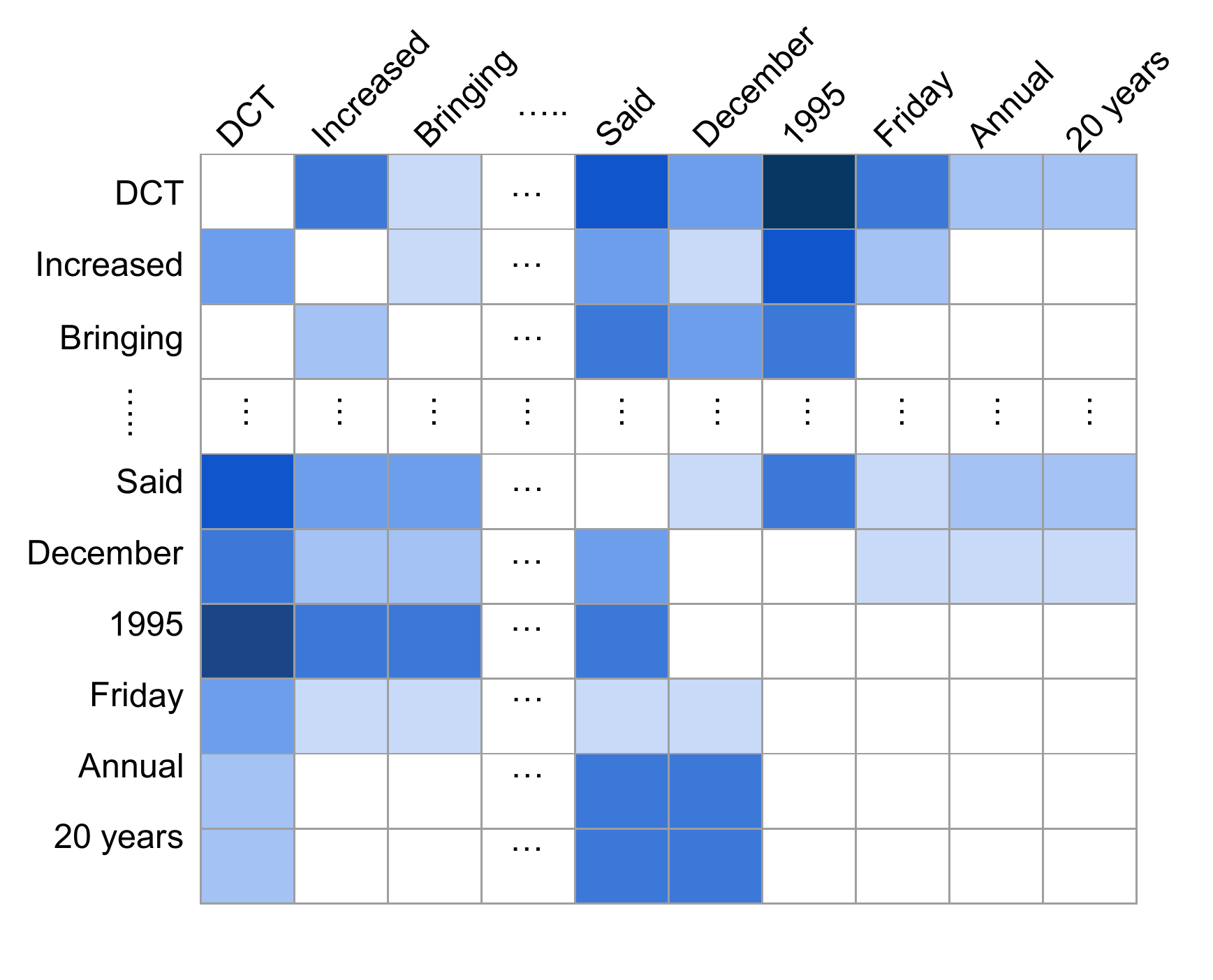}
	\caption{\label{fig:DCT_Attention} Visualization of the average edge attention of the temporal graph as learnt by \methodes{} for the document shown in \reffig{fig:Context_Attention}. Brighter colour implies higher attention. The correct DCT is 1996.
	}
\end{figure}

\begin{table}[t]
	\centering
	\begin{tabular}{lcc}
		\toprule
		Method 			 & Accuracy \\
		\midrule		
		\addlinespace
		T-GCN of NeuralDater			& 61.8  \\
		\methodes{}				 & \textbf{63.9} \\
		\midrule		
		\addlinespace
		S-GCN of NeuralDater          &63.2\\
		\methodcs{} 				& \textbf{65.6}\\
		\bottomrule
	\end{tabular}
	\caption{\label{tb:atten_result}Accuracy (\%) comparisons of component models with and without Attention (on APW dataset). This results show the effectiveness of both word attention and attentive graph convolution for this task. Please see \refsec{sec:atten_perf} for more details.}
\end{table}
Attentive Graph Convolution (\refsec{sec:graph_atten}) proves to be effective for \methodes{}, giving a 2\% accuracy improvement over non-attentive T-GCN of NeuralDater (Table \ref{tb:atten_result}). Similarly the efficacy of word level attention is also prominent from \reftbl{tb:atten_result}.

We have also analyzed our models by visualizing attentions over words and attention over graph nodes. Figure \ref{fig:Context_Attention} shows that \methodc{} focuses on temporally informative words or time mentions like ``1995'', alongside important contextual words like ``inflation'', ``Israel'' etc. For \methodes{}, from \reffig{fig:DCT_Attention} we observe that ``DCT'' and time-mention ``1995'' grabs the highest attention. Attention between ``DCT'' and other event verbs indicating past tense are quite prominent, which helps the model to infer 1996 (which is correct) as the most likely time-stamp of the document. These analyses provide us with a good justification for the performance of our attentive models.

\subsubsection{\textbf{Discussion and Error Analysis}}
\label{sec:discussion}

\begin{figure}[t]
	\centering
	\includegraphics[width=3.3in]{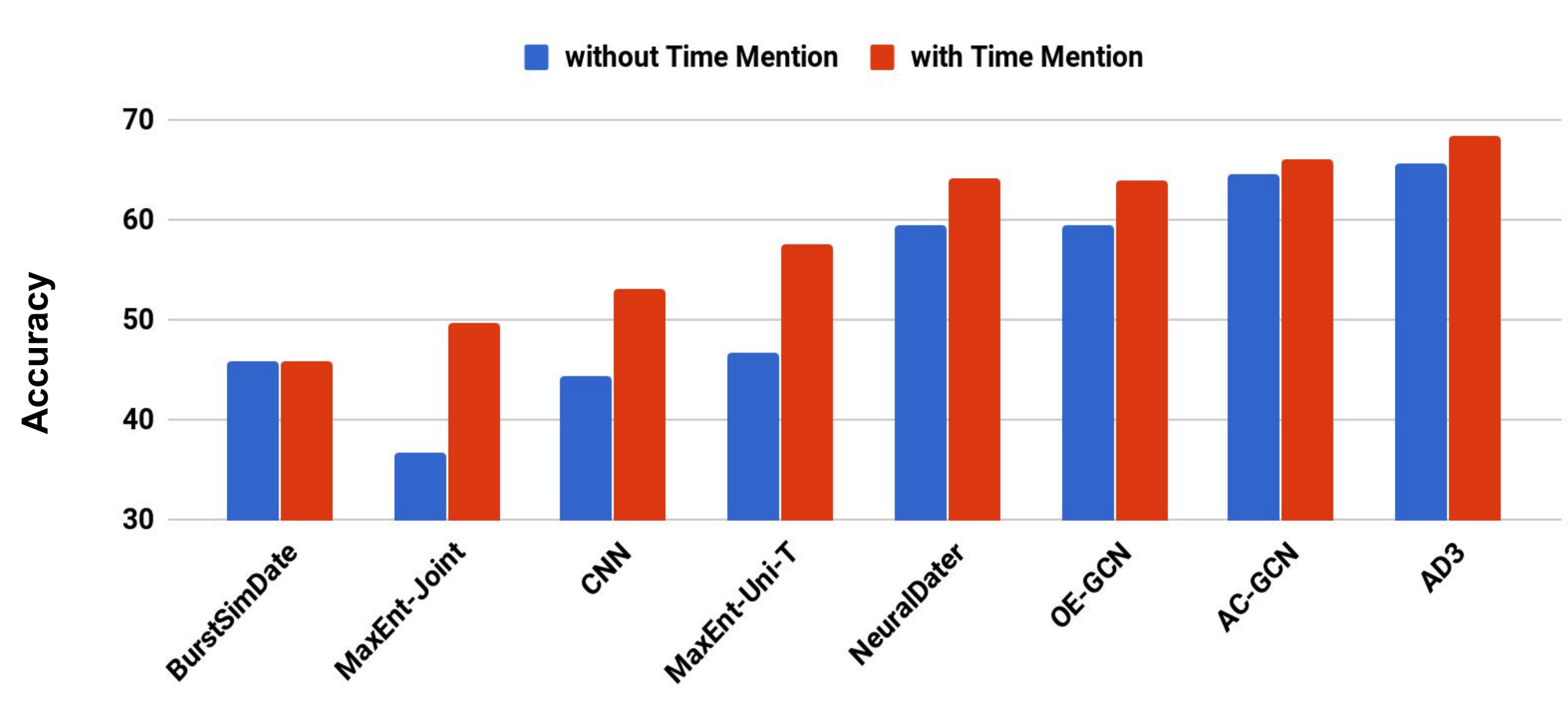}
	\caption{\label{fig:results_time_mention}Evaluating performance of different methods on dating  documents with and without time mentions (on APW dataset). Please see \refsec{sec:discussion} for details.}
\end{figure}

In this section, we list some of our observations while trying to identify pros and cons of our proposed methods. We divided the development split of the APW dataset into two sets -- those with and without any mention of time expressions (year). We apply our models and other methods to these two sets of documents and report accuracies in \reffig{fig:results_time_mention}. We find that overall, our methods performs better in comparison to the existing baselines in both scenarios. Even though the performance of our models degrades in the absence of time mentions, its performance is still the best relatively. 

\begin{figure}[t!]
	\centering
	\includegraphics[width=2.8in]{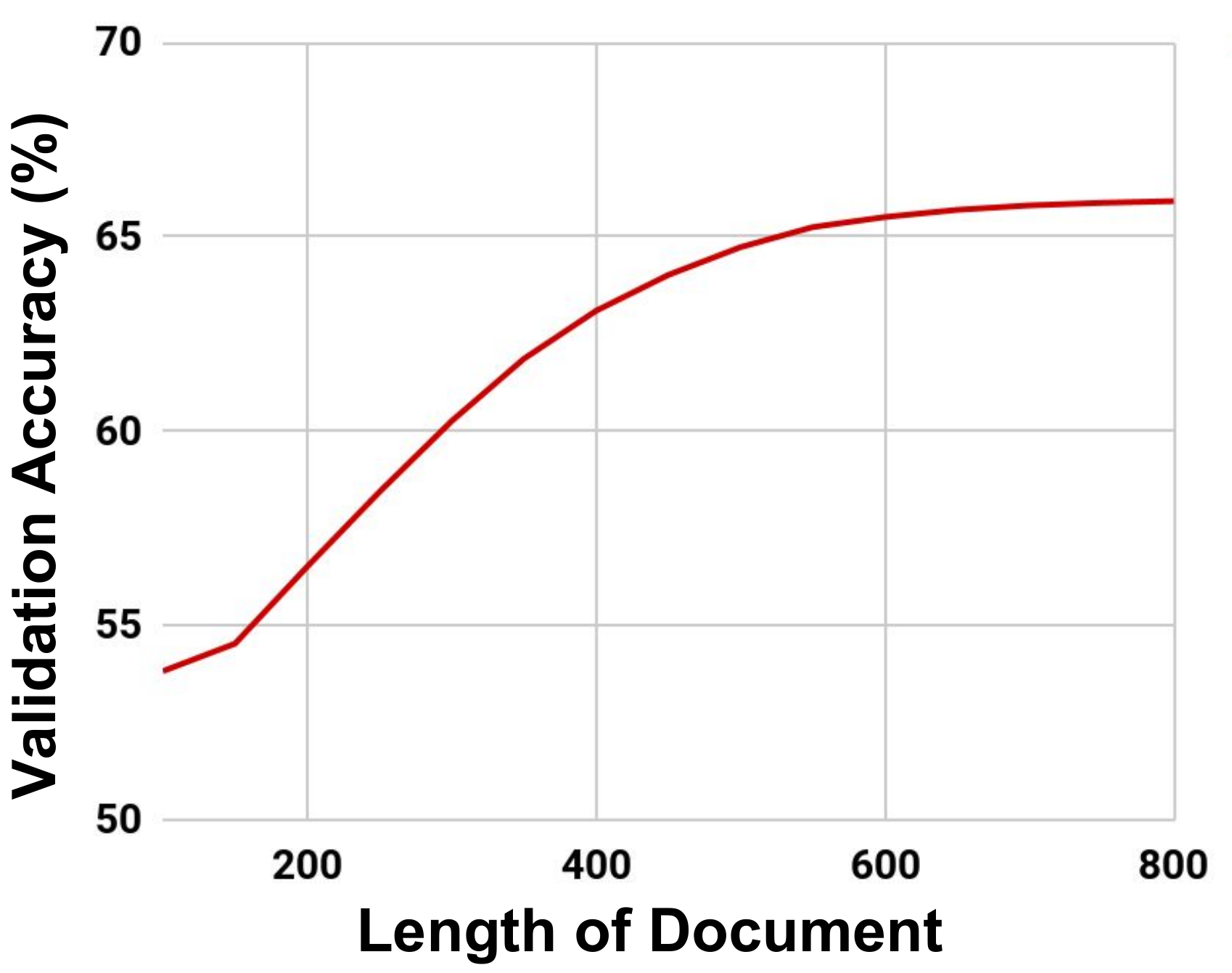}
	\caption{\label{fig:context_vs_length} Variation of validation accuracy (\%) with respect to length of documents (for APW dataset) for \methodcs{}. Documents having more than 100 tokens are selected for this analysis. Refer \refsec{sec:discussion}
	}
\end{figure}

\begin{figure}[t!]
	\centering
	\includegraphics[width=3in]{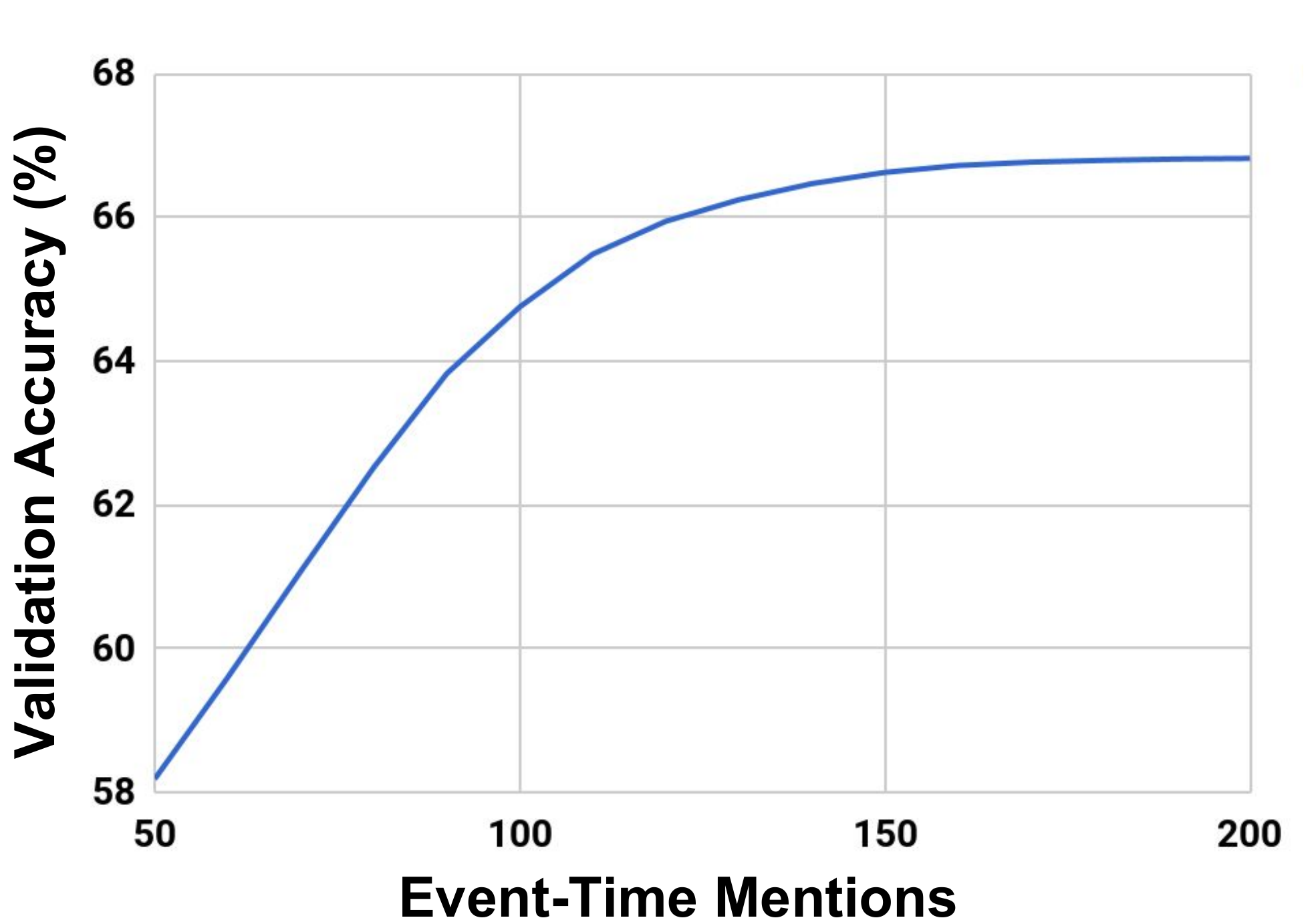}
	\caption{\label{fig:DCTpref_with_ET} Variation of validation accuracy (\%) with number of event-time mentions in documents (for APW dataset) for \methodes{}. Documents with more than 100 tokens are selected for this analysis. Refer \refsec{sec:discussion} 
	}
\end{figure}

Apart from empirical improvements over previous models, we also perform a qualitative analysis of the individual models - \methodes{} and \methodcs{}. Figure \ref{fig:context_vs_length} shows that the performance of \methodcs{} improves with the length of documents, thus indicating that richer context leads to better model prediction.  \reffig{fig:DCTpref_with_ET} shows how the performance of \methodes{} improves with the number of event-time mentions in the document, thus further reinforcing our claim that more temporal information improves model performance.

 Based on other analysis we find that \method{} becomes confused in the presence of multiple misleading time mentions; it also loses out on documents discussing events which are outside the time range of the text on which the model was trained. We have successfully tied up these loose ends with \methodfin{}, using attentive graph convolution, which successfully filters out noisy time mentions as is evident from \reffig{fig:DCTpref_with_ET}. But still our methods fail to identify timestamps of documents reporting local infrequent incidents without explicit time mention.



\subsection{\textbf{Conclusion and Future Work}}
\label{sec:conclusion}

We propose \method{} and \methodfin{}, which exploits both syntactic and temporal information in documents in a principled manner. To the best of our knowledge, this is the first application of deep learning, or more specifically attention based deep learning for dating documents. Our experimental results demonstrate the effectiveness of our models over all previous models. We also visualize the attention weights to show that the attentive models are able to choose what is important for the task and filter out noise, inherent in natural language. We are hopeful that the representation learning techniques explored in this paper will inspire further development and adoption of such techniques in the temporal information processing research community. We believe further research is required to successfully solve this problem and towards that goal we would like to incorporate external knowledge as a side information for time-stamping documents. In future, we plan to incorporate additional signals from Knowledge Graphs about entities mentioned in the document. We also plan to utilize free text temporal expression \cite{temponym_paper} in documents for improving performance on this problem.
\section{\textbf{Temporal Knowledge Graph Embedding:}}
\label{sec:Knowledge}
\subsection{\textbf{Introduction}}
\label{sec:introduction}

{\bf{Temporal Knowledge Graph Embedding:}}
Knowledge Graphs (KGs) are large multi-relational graphs where nodes correspond to entities, and edges represent relationships among them. Examples of a few KGs include NELL \cite{nell}, YAGO \cite{yago}, and Freebase \cite{freebase}. KGs have been found to be useful for a variety of tasks, viz., Information Retrieval \cite{tapping_kb_search,query_expansion_freebase}, Question Answering \cite{q&a_freebase,q&a_memorynet, q&a_2}, among others.

KG embedding has emerged as a very active area of research over the last few years, resulting in the development of several techniques \cite{transe,HoLE,distmult,transR,complex,conve}. These methods learn high-dimensional vectorial representations for nodes and relations in the KG,  while preserving various graph and knowledge constraints.

We note that KG beliefs are not universally true, as they tend to be valid only in a specific time period. For example, \textit{(Bill Clinton, presidentOf, USA)} was true only from 1993 to 2001. KG beliefs with such temporal validity marked are called as temporally scoped. These temporal scopes are increasingly available on several large KGs, e.g., YAGO \cite{yago}, Wikidata \cite{wikidata}. The mainstream KG embedding methods ignore the availability or importance of such temporal scopes while learning embeddings of nodes and relations in the KGs. These methods treat the KG as a static graph with the assumption that the beliefs contained in them are universally true. This is clearly inadequate and it is quite conceivable that incorporating temporal scopes during representation learning is likely to yield better KG embeddings.
In spite of its importance, temporally aware KG embeddings is a relatively unexplored area. Recently, a KG embedding method which utilizes temporal scopes was proposed in \cite{time_aware_link_pred}. However, instead of directly incorporating time in the learned embeddings, the method proposed in \cite{time_aware_link_pred} first learns temporal order among relations (e.g., \textit{wasBorIn} $\rightarrow$ \textit{wonPrize} $\rightarrow$ \textit{diedIn}). These relation orders are then incorporated as constraints during the KG embedding stage. Thus, the embedding learned by \cite{time_aware_link_pred} is not explicitly temporally aware.

In order to overcome this challenge, in this paper, we propose \emph{Hyperplane-based Temporally aware KG Embedding (\methodkg{})}, a novel KG embedding technique which directly incorporates temporal information in the learned embeddings. \methodkg{} fragments a temporally-scoped input KG into multiple static subgraphs with each subgraph corresponding to a timestamp. \methodkg{} then projects the entities and the relations of each subgraph onto timestamp specific hyperplanes. We learn the hyperplane (normal) vectors and the representation of the KG elements distributed over time jointly. Our contributions for this task are as follows:

\begin{itemize}
	\item We draw attention to the important but relatively unexplored problem of temporally aware Knowledge Graph (KG) embedding. In particular, we propose \methodkg{}, a temporally aware method for learning Knowledge Graph (KG) embedding.   
	\item In contrast to previous time-sensitive KG embedding methods, \methodkg{} encodes temporal information directly in the learned embeddings.
	\item Through extensive experiments on multiple real-world datasets, we demonstrate \methodkg{}'s effectiveness.
\end{itemize}

\subsection{\textbf{Related Work}}
\label{sec:related_work}

{\bf Temporal fact}:

Time, apart from being an information, also introduces a separate dimension to knowledge. Thus temporal scoping of relational facts is an imperative part of automatic knowledge graph construction and completion. T-YAGO \cite{timely_yago} extracts temporal facts from semi-structured data like Wikipedia, Infoboxes, and categories using only regular expressions. On the other hand, systems like PRAVDA  harvests temporal information from free text sources using label propagation. CoTS \cite{coupled_Temporal_learning} uses integer linear program based approach to model temporal constraints and proposes joint inference framework with few seed examples.

The task of extracting temporally rich events and time expressions and ordering between them is introduced in TempEval challenge \cite{tempeval13,tempeval10}. Various approaches \cite{caveo_plus,catena_paper} made for solving the task proved to be effective in other temporal reasoning tasks. Although we try to attend similar problem, the method proposed in this paper is more related to relational embedding learning paradigm than scoping temporal facts from  the web.

{\bf Relational Embedding learning methods}:

An enormous amount of research has been done in this field, especially for KG completion or link prediction task \cite{transe}. \cite{kg_review} provides a detailed review of the recent KG embedding learning methods. These can be broadly categorized into two different paradigms. TransE\cite{transe}, TransH\cite{transH}, TransR \cite{transR}, TransD \cite{transD} are the translational distance based models. Here the main theme is to minimize the distance between two entity vectors where one of them is translated by a relation vector. The realm of matrix factorization based methods includes bilinear model RESCAL \cite{Rescal}, DistMult \cite{distmult}, HoIE \cite{HoLE}. Some of the other notable models are Neural Tensor Networks(NTN) \cite{ntn}. We also provide some background on the traditional methods in section \ref{sec:background}. However, temporal dimension remains silent in all of these inference methods.

Link prediction through embeddings of the graph nodes and edges, are not only useful for inference over KG but also important for predicting incomplete pieces of the KG itself. Learning temporally steered embeddings is an important but proportionately less explored problem. Only some handful of methods have been proposed for this purpose. t-TransE \cite{time_aware_link_pred} learns time aware embedding by learning relation ordering jointly with TransE. They try to inflict temporal order on time-sensitive relations e.g. $ wasBornIn \rightarrow wonPrize \rightarrow diedIn$. t-TransE does not use the time information directly, whereas we incorporate time directly in our learning algorithm. Another approach Know-Evolve\cite{knowevolve} models the non-linear temporal evolution of KG elements using bilinear embedding learning method. They deploy recurrent neural network to capture non-linear dynamical characteristics of the embeddings. However, they restrict their domain to event-based interaction type of datasets which are fairly dense in nature.  \cite{time_validity} propose a method for temporal embedding learning using side information from the atemporal part of the graph. However, we use purely temporal KG to learn the temporally aware embedding.

\subsection{\textbf{Background: KG Embedding}}
\label{sec:background}


\begin{figure}[t]
	\centering
	\includegraphics[scale = 0.2]{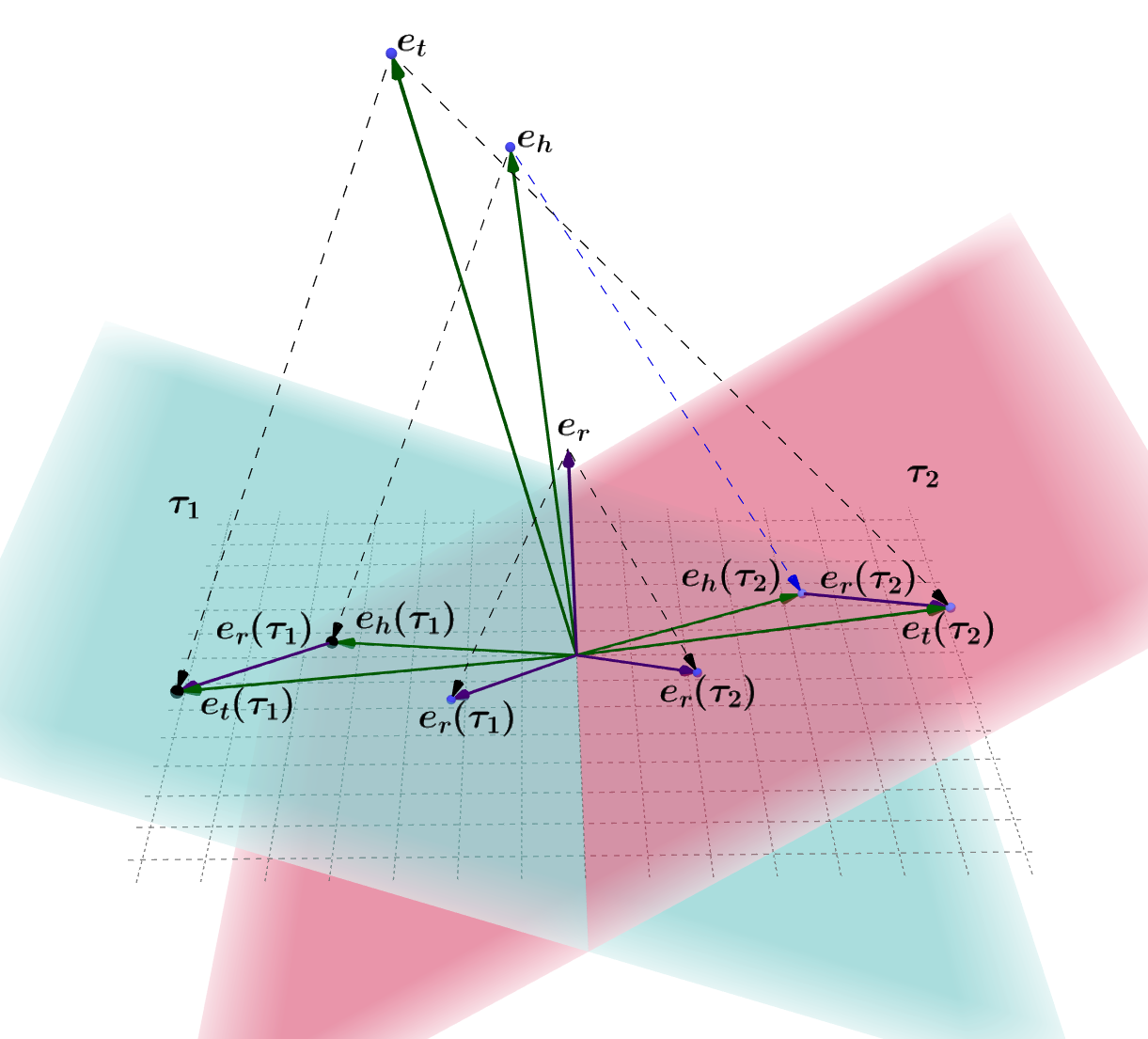}
	\caption{\label{fig:overview_kg} In the figure, the vectors $e_{h}, e_{r} \text{ and } e_{t}$  correspond to the triple $(h,r,t)$ that is valid at time $\tau_1$ and $\tau_2$ respectively. $e_{h}(\tau_{1}), e_{r}(\tau_{1}) \text{ and } e_{t}(\tau_{1})$ are the projections of this triple on the hyperplane corresponding to time $\tau_{1}$ (similarly for time $\tau_{2}$). Our method \methodkg{} minimizes the  translational distance, $\sum_{i}\Vert e_{t}(\tau_{i})+e_{r}(\tau_{i}) - e_{t}(\tau_{i})  \Vert_{1}$  in order to learn the temporal representations of entities and relations for this triple.}	
\end{figure}

In this section, we provide an overview of the existing methods for knowledge graph representation learning \cite{transe}, \cite{transH}. Consider a KG $\m{G}$ with a set of entities $\m{E}$. The set of directed edges, $\m{D}^{+}$ consists of triples $(h,r,t)$, where the edge direction is from $h$ to $t$ and the edge label (also popularly known as relation) is $r$.


\subsubsection{\textbf{TransE and TransH}}
\label{sec:TransE}
TransE \cite{transe} is a simple and efficient translational distance model. It interprets the relation as a translation vector between head and tail entity vectors. Given two entity vectors $e_{h}$, $e_{t}$ $\in \mathbb{R}^{n}$, it tries to map the relation as a translation vector $e_{r}\in \mathbb{R}^{n}$ i.e. $e_{h}+ e_{r}  \approx e_{t}$ for observed triple $(h,r,t)$. So the distance based scoring function used for plausible triples is hereby , 
$$f(h,r,t)=\Vert e_{h}+e_{r}-e_{t}\Vert_{l_{1}/l_{2} },$$
where, $\Vert \cdot \Vert_{l_{1}/l_{2} }$ is the $l_{1}$ or $l_{2}$-norm of the difference vector. $f(h,r,t)$ will be minimized for observed or correct triples. In order to differentiate between correct and incorrect triples, their TransE score difference is minimized using margin based pairwise ranking loss. More formally, we optimize 
$$ \sum_{x \in \m{D}^{+}}\sum_{y\in \m{D}^{-}}\max(0,f(x)-f(y)+\gamma),$$
with respect to the entity and relation vectors. $\gamma$ is a margin separating correct and incorrect triples. $\m{D}^{+}$ is the set of all positive triples i.e. observed triples in KG. The negative samples are drawn randomly from the set
\begin{multline*}
\m{D}^{-}=  \{(h^{'},r,t)|h^{'}\in \m{E}, (h^{'},r,t)\notin \m{D}^{+} \}  \\
\cup \{(h,r,t^{'})|t^{'}\in \m{E}, (h,r,t^{'})\notin \m{D}^{+} \}.
\end{multline*}
TransE fails to model the many-to-one, one-to-many, many-to-many type of relations as it does not learn a distributed representation of entities when it is involved with many relations. To tackle these situations, TransH was proposed. TransH \cite{transH} models a relation $r$ as a vector on a relation specific hyperplane and project entities associated with it on that particular hyperplane in order to learn distributed representation of the entities.

We notice that not only the role of the entities changes with time but also the relationship between them changes. Taking some inspiration from the objective of \cite{transH}, we propose a hyperplane based method for learning KG representation distributed in time.
\subsection{\textbf{Proposed Method: \methodkg{}}}
\label{sec:model}
In this section, we present a detailed description of \methodkg{} (\reffig{fig:overview_kg}) which not only exploits the relational properties among entities but also uses the temporal meta-data associated with them. 

\subsubsection{\textbf{Temporal Knowledge Graph}}
\label{sec:temporal_graph}

Usually knowledge graphs are treated as a static graph consisting of triples in form of $(h,r,t)$. Adding a separate time dimension to the triple makes the KG dynamic. Consider the quadruple $(h,r,t, [\tau_{s}, \tau_{e}])$, where $\tau_{s}$ and $\tau_{e}$ denote the start and end time during which the triple $(h,r,t)$ is valid. Unlike \cite{time_aware_link_pred}, we incorporate this time meta-facts directly into our learning algorithm to learn temporal embeddings of the KG elements. Given the timestamps, the graph can be dismantled into several static graphs consisting of triples that are valid in the respective time steps, e.g., knowledge graph $G$ can be expressed as $\mathbb{G} = \mathbb{G}_{\tau_{1}} \cup \mathbb{G}_{\tau_{2}}\cup  \cdots \cup  \mathbb{G}_{\tau_{T}}$ , where $\tau_{i}$, $i\in 1,2, \cdots, T $ are the discrete time points.\\
We constructed this temporal component-graphs ($\mathbb{G}_{\tau}$) from the quadruples by considering $(h,r,t)$ to be a positive triple at each time point between $\tau_{s}$ and $\tau_{e}$.
Now, given a quadruple $(h,r,t, [\tau_{s},\tau_{e}])$, we consider it to be a positive triple for each time point between $\tau_{s}$ and $\tau_{e}$. So, we include $(h,r,t)$ in each $\mathbb{G}_{\tau}$ , where $\tau_{s}\leq\tau\leq\tau_{e}$. The set of positive triple corresponding to time $\tau$ is denoted as  $\m{D}_{\tau}^{+}$.

\subsubsection{\textbf{Projected-Time Translation}}
\label{sec:time_proj}

TransE considers entity and relation vectors in the same semantic space for a static graph. We observe that time is the main source of different many-to-one, one-to-many or many-to-many relations, e.g. $(h,r)$ pair can be associated with different tail entity $t$ at different points of time. Thus traditional methods fail to disambiguate them directly. In our time guided model, we want the entity to have a distributed representation associated with different time points.

We represent time as a hyperplane ie. for T number of time steps in the KG, we will have T different hyperplanes represented by normal vectors $w_{t_{1}},w_{t_{2}}, \cdots, w_{t_{T}} $. Thus we try to segregate the space into different time zones with the help of the hyperplanes. Now, triples valid at time $\tau$ (i.e. the sub graph $\mathbb{G}_{\tau} $) are projected onto time specific hyperplane $w_{\tau}$, where their translational distance (TransE  (\refsec{sec:background} our case) is minimized. To illustrate, form \reffig{fig:overview_kg} the triple $(h,r,t)$ is valid for both time frame $\tau_{1} \text{ and } \tau_{2}$. Hence they are projected on hyperplanes corresponding to those times.\\
Now we compute the projected representation on $w_{\tau}$ as, 
$$P_{\tau}(e_{h})  = e_{h} - (w_{\tau}^\top e_{h})w_{\tau}, $$
$$P_{\tau}(e_{t})   = e_{t} - (w_{\tau}^\top e_{t})w_{\tau}, $$
$$P_{\tau}(e_{r})   = e_{r} - (w_{\tau}^\top e_{r})w_{\tau}, $$
where we restrict  $\Vert w_{\tau}	\Vert_{2} = 1$.

We expect that  a positive triple, valid at time $\tau$ will have the mapping as $P_{\tau}(e_{h}) + P_{\tau}(e_{r}) \approx P_{\tau}(e_{t})$, thus we use the scoring function,
$$f_{\tau}(h,r,t)=\Vert P_{\tau}(e_{h})+P_{\tau}(e_{r}) - P_{\tau}(e_{t})  \Vert_{l_{1}/l_{2} }.$$

We learn $\{w_{\tau}\}^{T}_{\tau=1}$ for each time stamp $\tau$, along with the entity and relation embeddings. So, by projecting the triple into its time hyperplane we incorporate temporal knowledge into the relation and entity embeddings i.e. the same distributed representation will have a different role in different points of time.\\
\textbf{Optimization : } As mentioned in section \ref{sec:TransE} , we use the margin-based ranking loss:
$$\m{L} = \sum_{\tau \in [T]}\sum_{x \in \m{D}_{\tau}^{+}}\sum_{y \in \m{D}_{\tau}^{-}}\max(0,f_{\tau}(x)-f_{\tau}(y)+\gamma),$$
where,  $\m{D}_{\tau}^{+}$ is the set of valid triples with time-stamp $\tau$. The negative samples are drawn from the set of all negative samples, $\m{D}_{\tau}^{-}$ which considers the set of all the triples that does not belong to the KG, irrespective of timestamps.  More formally, for time step $\tau$ the negative samples are drawn from the set:

	\begin{multline}
	\m{D}_{\tau}^{-}=  \{(h^{'},r,t,\tau)|\text{ }h^{'}\in \mathcal{E}, (h^{'},r,t)\notin \m{D}^{+} \}  \\
	\cup \{(h,r,t^{'},\tau)|\text{ }t^{'}\in \mathcal{E}, (h,r,t^{'})\notin  \m{D}^{+}  \}.
	\label{eqn:tans}
	\end{multline}

The above mentioned loss $\m{L}$ is minimized subjected to the constrains.
\[
\Vert e_{p} \Vert_{2} \leq 1 , \forall \ p \in \m{E},~~~\Vert w_{\tau}	\Vert_{2} = 1, \forall \ \tau \in [T]
\]
We enforce the first one by adding $l_{2}$- regularization of entity vectors with $\m{L}$. We take care of the second constraint by normalizing the time embeddings viz. the hyperplane normal vectors after each update of stochastic gradient descent.

\subsection{\textbf{Experimental Setup}}
\label{sec:experiments}

We evaluate our model and compare with different state-of-the-art baselines based on Link prediction (\refsec{sec:entity_prediction}) task. Evaluation metrics used are same as that of the traditional KG embedding method \cite{transe} for link prediction task.

\subsubsection{\textbf{Datasets}}
\label{sec:data}
Knowledge Graphs such as Wikidata \cite{wikidata} and YAGO \cite{yago} have time annotations on a subset of the facts. We extracted the temporally rich subgraph from them for testing our algorithm as well as the baselines.\\
\textbf{YAGO11k:} In the knowledge graph YAGO3 \cite{yago3}, some temporally associated facts have meta-facts as \textit{(\#factID, occurSince, $\textit{t}_{\textit{s}}$)}, \textit{(\#factID, occurUntil, $\textit{t}_{\textit{e}}$)}. The total number of time annotated facts containing both occursSince and occursUntil are 722494. Out of them, we selected top 10 most frequent temporally rich relations. In order to handle, sparsity we recursively remove edges containing entity with only a single mention in the subgraph. This ensures a healthy connectivity within the graph. Finally, we obtain a purely temporal graph of 20.5k triples and 10623 entities by following this procedure.\\
\textbf{Wikidata12k}: We extracted this temporal Knowledge Graph from a preprocessed dataset of Wikidata proposed by \cite{time_validity}\footnote{https://staff.aist.go.jp/julien.leblay/datasets/}. We followed a similar procedure as described in YAGO11k. Here also, we distill out the subgraph with time mentions for both start and end. We ensure that no entity has only a single edge connected to it. We select top 24 frequent temporally rich relations for this case, which resulted in 40k triples with 12.5k entities. The dataset is almost double in size with respect to YAGO11k, which again reinforces the effectiveness of our model with higher confidence.
\begin{table}[t]
	\centering
	\begin{small}
		\begin{tabular}{cccc}
			\toprule
			Datasets 	& \# Entity &  \#relations &Train/Valid/Test \\
			\midrule
			Wikidata12K 		&  12,554	&  24 & 32,497/4,062/4,062\\
			YAGO11K		&  10,623	& 10  & 16,408/2,050/2,051\\
			\bottomrule
			\addlinespace
		\end{tabular}
		\caption{\label{tb:datasets}Details of datasets used. Please see \refsec{sec:data} for details.}
	\end{small}
\end{table}
\begin{table*}[t!]
	\begin{small}
		\begin{center}
			\begin{tabular}{ |c|cc|cc|cc|cc|}
				\hline \multicolumn{1}{|c|}{Dataset}  & \multicolumn{4}{|c|}{YAGO11K} & \multicolumn{4}{|c|}{Wikidata12K} \\ 
				\hline 
				\multirow{2}{*}{Metric} & \multicolumn{2}{|c|}{Mean Rank} & \multicolumn{2}{|c|}{Hits@10(\%)} &\multicolumn{2}{|c|}{Mean Rank} & \multicolumn{2}{|c|}{Hits@10(\%)} \\
				& tail & head & tail & head & tail & head &  tail & head\\
				\hline 
				Trans-E & 504  & 2020   & 4.4       &  1.2      &  520 & 740  & 11.0        & 6.0\\ 
				TransH\cite{transH}& 354 & 1808 &5.8 &1.5 & 423 &648    &23.7 & 11.8\\ 
				HolE\cite{HoLE}& 1828 & 1953 & 29.4 & 13.7 & 734& 808& 25.0 &12.3\\ 
				t-TransE \cite{time_aware_link_pred}& 292  & 1692  &6.2 &1.3 & 283 & 413 & 24.5 &14.5\\
				\hline 
				\textbf{\methodkg} & \textbf{107} & \textbf{1069} & \textbf{38.4} & \textbf{16.0} & \textbf{179} & \textbf{237} & \textbf{41.6} & \textbf{25.0}\\
				\hline
			\end{tabular}
			\caption{\label{tb:link_pred_result} Mean Rank (lower the better) and Hits@10(higher the better) for different methods for Link Prediction task. Proposed method \methodkg{} outperforms all the traditional approaches. The better performance of \methodkg{} over t-TransE can be attributed to the fact that it incorporates time directly. Please see \refsec{sec: model_perp} for details.}
		\end{center}
	\end{small}
\end{table*}

%


\subsubsection{\textbf{Baselines}} For evaluating  the performance of our algorithm, we compare against the following methods:

\begin{itemize}
	\item \textbf{t-TransE} \cite{time_aware_link_pred}: This method uses a temporal ordering of relations to model knowledge evolution in temporal dimension. They regularize the traditional embedding score function with observed relation ordering with respect to head entities. 
	\item \textbf{HolE} \cite{HoLE}: We take this method as a representative of state-of-the-art KG representation learning method and  demonstrate that time guided TransE can outperform this method using side information of time meta-facts in principled manner.
	\item \textbf{TransE} \cite{transe},  \textbf{TransH} \cite{transH}: These are the simple distance based model. We build  \methodkg{} on top of TransE and demonstrate the gains over these methods.

\end{itemize}
\subsubsection{\textbf{Entity Prediction}}
\label{sec:entity_prediction}
The task is to predict the missing entity, given an incomplete relational fact with its time. We experimented with both YAGO11K and Wikidata12k dataset. Training is done in perspective of both head and tail prediction. More formally, for generation of negative sample from a correct triple $(h,r,t,\tau)$, we split them in two parts - $(h,r,?,\tau)$ (for tail entity prediction) and $(?,r,t,\tau)$.(for head entity prediction). In this task we follow \refeqn{eqn:tans} for generating negative samples i.e. for each of tail and head query terms we randomly replace an entity such that newly generated triple is not observed in the graph, for eg, we sample $t'$ such that $t'\in \mathscr{E}\setminus t $ and $(h,r,t^{'},\tau)\notin D_{\tau}^{+}$.\\

\textbf{Ranking Protocol}: For a test triple $(h,r,t,\tau)$, we generate corrupted triples by replacing tail entity (for tail prediction) or head entity (for head prediction) with all possible entities. Filtered protocol, proposed by \cite{transe} says that the corrupted triples must not be a part of the graph itself. To illustrate, given a test triple $(h,r,t)$ for tail prediction task, we compute scores for the candidate set $C(h,r) = \{(h,r,t^{'}: \forall t^{'}\in \mathbb{E})\}\setminus(Train\cup Test\cup Valid) \cup {(h,r,t)}$ . We rank all the triples in $C(h,r)$ in the increasing order of their score and find the rank of the actual triple $(h,r,t)$. We report the mean rank over all the test queries (MR) and proportion of correct entities in top 10 rank (Hits@10).



\textbf{Implementation Details:} 
For all the methods, we have kept batch size b = 50k on both the datasets. The dimensions of the embeddings ($d$) are varied in the range \{64, 128, 256\}. The margins($\eta$) for all the methods are chosen from the set \{1, 2, 5, 10\}. Learning rate used for SGD, $lr \in$ \{0.01, 0.001, 0.0001\}.\\
The best configuration is chosen by corresponding lowest MR on the validation set. For both YAGO11k and Wikidata12k, we obtained $d$ = 128, $\eta$ = 10, $lr$ = 0.0001 using $\mathrm{l}_{1}$-norm in the scoring function.\\
Both YAGO11k and Wikidata12k contain time annotations to the granularity of days. We only deal with year level granularity by dropping the month and date information. Timestamps are then treated as 61 different intervals for YAGO and 78 different intervals for Wikidata. The main motive behind having time classes is to distribute the time annotations in the KG uniformly. For example, less frequent year mentions are clubbed into same time class but years with high frequency forms individual classes. To illustrate, in Wikidata there are classes like 1596-1777, 1791-1815 with a large span as the events occurring on those points of time are quite less in KG. The years like 2013, 2014 being highly frequent are self-contained.

\subsubsection{\textbf{Performance analysis \& comparison}}
\label{sec: model_perp}
The obtained results for the task is based on the above mentioned hyperparameters.

The results reported in Table \ref{tb:link_pred_result} demonstrate the efficacy of \methodkg{}. We observe that our model outperforms the traditional state-of-the-art link prediction model HolE \cite{HoLE} by a significant margin in both the datasets. We also show a large boost in performance over TransE \cite{transe}. This significant gain empirically validates our claim that including temporal information in a principled fashion helps to learn richer embeddings of the KG elements. We notice that HolE is performing significantly poor in terms of MR but it exceeds the other baselines in Hits@10 by a large margin.  \\
Again, in comparison with the temporal model t-TransE \cite{time_aware_link_pred}, \methodkg{} proves to be effective. t-TransE performs better than TransE and HolE due to its implicit time incorporation through relation ordering. \methodkg{} with its direct inclusion of time in the relation-entity semantic space outperforms all of them.  \\

\subsection{\textbf{Conclusion and Future Work}}
\label{sec:conclusion}

We propose \methodkg{}, a hyperplane-based method for learning knowledge graph embeddings. We also extract a purely temporal dataset from the practical real-world KGs and demonstrate the effectiveness of our model over both traditional and time aware embedding methods for temporal link prediction. In future, we would like to incorporate type consistency information to further improve our model. We are hopeful that our proposed temporal representation learning algorithm will further motivates research towards temporal KG embedding learning.

\section{\textbf{Acknowledgements:}}

I would like to thank my advisor, Dr. Partha Pratim Talukdar, for his patience, guidance and immense support throughout the last one year. I feel extremely fortunate and grateful to have him as my advisor. The document dating method, \method{}, is based on \cite{NeuralDater} which is a published paper in ACL, 2018 with Shikhar Vasisth and Shib Sankar Dasgupta as co-authors. The attention based document dating system \methodfin{} and the temporally aware KG embedding, \methodkg{}, are based on submitted papers in EMNLP, 2018 where Shib Sankar Dasgupta is a co-author. I would like to thank all my co-authors; without them the project would not have been completed.

 I thank all my MALL Lab mates for their companionship and helpful advices and making the journey enjoyable. I am grateful to the Department of Computational and Data Sciences and MALL Lab for providing me with such high quality research facility. Last but not least, I would like to express my heartfelt gratitude towards my parents, brothers, uncles and aunts and Amrita Dasgupta for their continued support.

%
%
%
%
%
%
%
%
%
%
%
%

\bibliographystyle{IEEEtran}
\bibliography{template}

\end{document}